\documentclass[manuscript,nonacm, screen]{acmart}
\AtBeginDocument{%
  \providecommand\BibTeX{{%
    \normalfont B\kern-0.5em{\scshape i\kern-0.25em b}\kern-0.8em\TeX}}}

\usepackage{caption}
\definecolor{todoblue}{RGB}{0, 91, 187}

\setcopyright{acmcopyright}
\copyrightyear{2023}
\acmYear{2023}
\acmDOI{XXXXXXX.XXXXXXX}

\acmConference[FAccT 2023]{Make sure to enter the correct
  conference title from your rights confirmation email}{July 12--15,
  2023}{Chicago, IL}
\acmBooktitle{2022 ACM Conference on Fairness, Accountability, and Transparency (FAccT '22), June 12--15, 2023, Chicago, IL} 
\acmPrice{15.00}
\acmISBN{978-1-4503-XXXX-X/18/06}

\begin{document}

\title[More Data Types More Problems]{More Data Types More Problems: A Temporal Analysis of Complexity, Stability, and Sensitivity in Privacy Policies}

\author{Juniper Lovato}
\orcid{0000-0002-1619-7552}
\affiliation{%
  \institution{Vermont Complex Systems Center, University of Vermont}
  \streetaddress{82 University Place}
  \city{Burlington}
  \state{Vermont}
  \country{USA}}
\email{juniper.lovato@uvm.edu}

\author{Philip Mueller}
\affiliation{%
  \institution{Vermont Complex Systems Center, University of Vermont}
  \streetaddress{82 University Place}
  \city{Burlington}
  \state{Vermont}
  \country{USA}}

\author{Parisa Suchdev}
\affiliation{%
  \institution{Computer Science, University of Vermont}
  \streetaddress{82 University Place}
  \city{Burlington}
  \country{USA}}

\author{Peter S. Dodds}
\affiliation{%
  \institution{Computer Science, University of Vermont}
  \streetaddress{82 University Place}
  \city{Burlington}
  \country{USA}}

\renewcommand{\shortauthors}{Lovato et al.}

\begin{abstract}
Collecting personally identifiable information (PII) on data subjects has become big business. Data brokers and data processors are part of a multi-billion-dollar industry that profits from collecting, buying, and selling consumer data. Yet there is little transparency in the data collection industry which makes it difficult to understand what types of data are being collected, used, and sold, and thus the risk to individual data subjects. In this study, we examine a large textual dataset of privacy policies from 1997-2019 in order to investigate the data collection activities of data brokers and data processors. We also develop an original lexicon of PII-related terms representing PII data types curated from legislative texts. This mesoscale analysis looks at privacy policies overtime on the word, topic, and network levels to understand the stability, complexity, and sensitivity of privacy policies over time. We find that (1) privacy legislation correlates with changes in stability and turbulence of PII data types in privacy policies; (2) the complexity of privacy policies decreases over time and becomes more regularized; (3) sensitivity rises over time and shows spikes that are correlated with events when new privacy legislation is introduced.
\end{abstract}

\begin{CCSXML}
<ccs2012>
   <concept>
       <concept_id>10003456.10003462.10003477</concept_id>
       <concept_desc>Social and professional topics~Privacy policies</concept_desc>
       <concept_significance>500</concept_significance>
       </concept>
   <concept>
       <concept_id>10003033.10003083.10003094</concept_id>
       <concept_desc>Networks~Network dynamics</concept_desc>
       <concept_significance>500</concept_significance>
       </concept>
   <concept>
       <concept_id>10002978.10003029.10011150</concept_id>
       <concept_desc>Security and privacy~Privacy protections</concept_desc>
       <concept_significance>500</concept_significance>
       </concept>
 </ccs2012>
\end{CCSXML}

\ccsdesc[500]{Social and professional topics~Privacy policies}
\ccsdesc[500]{Networks~Network dynamics}
\ccsdesc[500]{Security and privacy~Privacy protections}

\keywords{Privacy, NLP, Data Privacy, Data Ethics, Privacy Policies, Data Science, Networks}

\begin{teaserfigure}
  \centering
  \includegraphics[width=.5\columnwidth]{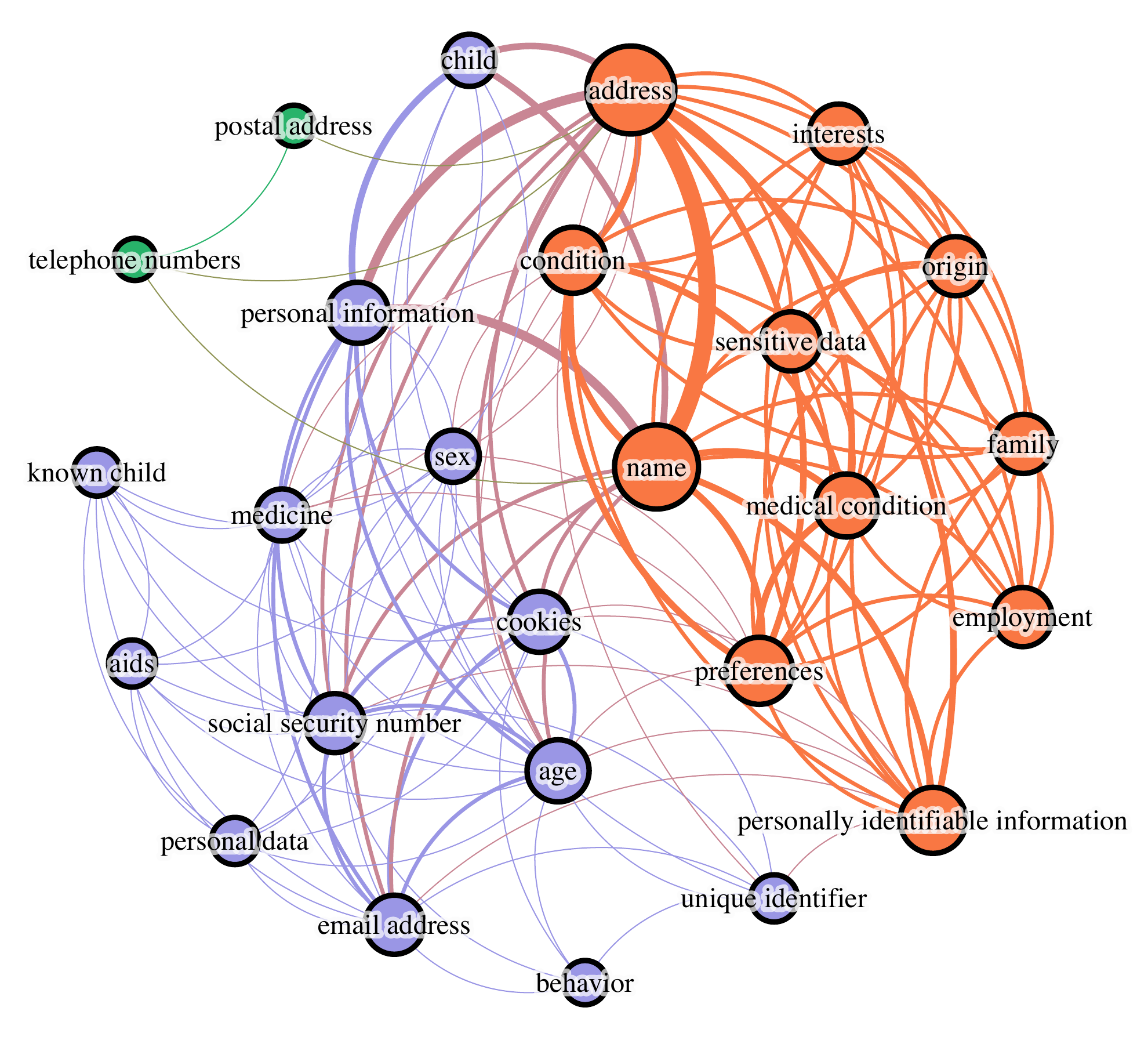}
  \caption{1997 co--occurrence network of PII relevant terms. All nodes with a degree less than one are filtered from the network (words that do not co--occur with other words). Networks are partitioned by modularity, signified by the nodes' color. The node size ranges from 30 to 60 based on the weighted degree.}
  \label{fig:teaser}
\end{teaserfigure}

% \received{30 January 2023}

\maketitle

\section{Introduction}
\label{section:introduction}

Data brokers and data processors (DBDPs) form a multi-billion-dollar industry that collects, buys, and sells personally identifiable information (PII) from individuals worldwide. According to the market research company eMarketer, the size of the data broker industry was approximately 300 billion dollars in 2020. They have also recently projected the industry's size to nearly double within three years. \cite{kirkpatrick2021monetizing}

Even with new legislation emerging in the U.S. and the E.U., the data broker industry lacks sufficient transparency and regulation. Moreover, many data processors may not be officially listed by definition as data brokers on any U.S. state registry, even though they play a significant role in collecting, processing, and sharing (often for monetary gain) data that they collect from consumers. It is important to note that many data processors do not sell raw data but rather generate income by serving as a data pass-through selling insights or using the data to generate income from advertising revenue. \cite{gonzalez2017fdvt} According to the Vermont Data Broker Act of 2018, a data broker is defined as ``a business, or unit or units of a business, separately or together, that knowingly collects and sells or licenses to third parties the brokered personal information of a consumer with whom the business does not have a direct relationship.'' \cite{lawvt2018} For example, Facebook may not be considered a data broker by definition. However, it is one of the largest data processors in the world and benefits financially from the data assets they collect from social media users. For the purposes of our project, we will consider both data brokers and data processors (we will collectively call them DBDPs) because both significantly impact the risks and potential harms to data subjects (individuals whose data is collected by DBDPs) associated with collecting personal data.

Lack of transparency and information about the DBDP industry \cite{crain2018limits} is a significant hurdle in assessing the harms associated with collecting PII. For our purposes, we are primarily concerned with harm to U.S. data subjects. 

There are several issues related to the lack of transparency in the DBDP industry, namely:  (1) It is difficult to properly assess the magnitude and impact of harm to data subjects because it is currently unknown how much data and what data types have been collected by DBDPs over time; (2) There is little to no mandatory financial reporting for the DBDP industry, so the market value of the data they collect and sell is unknown which makes it difficult to assess damages and loss; (3) Finally, it is difficult to track which third parties have been granted access to PII data by DBDPs and track the flow of information that is shared. 

Moreover, many DBDPs function under the justification that PII data is collected with sufficient consent from individuals. However, the legitimacy of this consent is a concern, particularly considering the necessary criteria for informed consent and the issue of group consent for data that implicates social groups. \cite{lovato2022limits} The legitimacy of individual informed consent is dependent on (1) the data subject understanding the agreement; (2) entering into it without coercion; (3) entering the agreement intentionally and deliberately; (4) and the agreement authorizing a specific course of action for the data being collected. \cite{faden1986history} If DBDPs were to follow the individual informed consent criteria, particularly for criteria 4, they would need a mechanism for the data subject to track the status of their data to assess if the consent agreement was being upheld. However, there is currently no way in the U.S. to track your PII data and fully understand how it is being processed and shared by DBDPs. Information asymmetry between DBDPs and data subjects makes it challenging for data subjects to follow their data, know what types of data are being collected, used, and shared, and hold firms accountable if data are being processed against the terms of their original consent agreement or the law. 

Typically, the only reporting required for DBDPs, in the United States includes declaring they are a data broker on a State registry (required by law in states like Vermont \cite{lawvt2018} and California \cite{lawca2018}) and publishing a privacy policy on their public website. Transparency will remain an issue until there is clear mandatory reporting by DBDPs on the data collected, used, and shared, a mechanism for consumers to track their data, and disclosure of detailed digital assets by DBDPs (currently, only publicly traded DBDPs need to disclose this information to the Securities and Exchange Commission and data assets are lumped together with all intangible assets like patents and intellectual property). In place of these resolutions, researchers must find other means to infer the activities of DBDPs. 

In this study, we perform a mesoscale exploratory data analysis on a temporal privacy policy dataset (includes years 1997-2019) and extract time series to explore the following measures: (1) descriptive summaries of the text such as basic summary statistics and topics, (2) frequency distributions of words as a measure of stability, (3) complexity as measured by corpus compression factor, (4) and co--occurrence network structures as a measure of sensitivity. 

We investigate text from privacy policies to infer DBDP activities and understand what PII data types  DBDPs collect and how this activity has changed over time. Our study looks at a temporal textual dataset from Amos et al. \cite{amos2021privacy} that includes over one million privacy policy snapshots from 1997-2019 from over 100,000 websites. We also use legislative text from eight U.S. state laws on data privacy to manually extract a lexicon of personally identifiable information (PII) terms represented in the corpora to represent PII data types. 

\begin{figure*}[ht]
    \centering
    \includegraphics[width=1\textwidth]{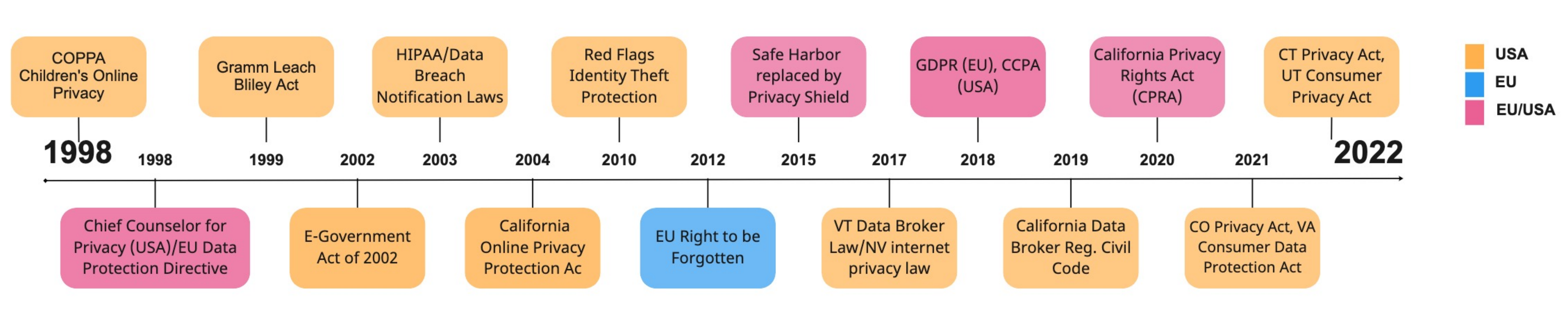}
    \caption{Timeline of selected privacy legislation in the U.S. and E.U. from 1998-2022}
    \label{fig:changes-timeline}
\end{figure*}

Finally, we compare signals in our results to privacy regulation events in the U.S. and the E.U. (we include some E.U. regulations events like GDPR because they have a global impact on privacy policy text) and find correlations between changes in privacy policies and new governmental regulations. A timeline of significant governmental regulations in the U.S. and E.U. from 1997-2022 can be seen in Figure \ref{fig:changes-timeline}. This multidisciplinary project uses methods from complex systems and data science, natural language processing (NLP), computational social science, and network science.

Near future work will examine the flow of these data by inferring the data types collected by DBDPs in the collection section of the text and then inferring the data types shared by DBDPs in the share section of the text. We will then investigate tort law to find examples of the magnitude of monetary harms associated with each relevant PII data types to estimate the potential monetary harm inflicted on data subjects through collecting and selling their PII data. 

This paper investigates four primary research questions: 

\textbf{Research Questions:} 

\begin{itemize}
    \item \textbf{Q1:} What PII-related words, topics, and co--occurrence network structures appear in privacy policies over time? 
    \item \textbf{Q2:} What PII data types are collected from consumers, and how stable is the representation of those PII data types over time? 
    \item \textbf{Q3:} How complex or regular are privacy policies over time? 
    \item \textbf{Q4:} What PII data types are collected concurrently from consumers, and how has the level of sensitivity (and potential risk to data subjects) changed over time? 
\end{itemize}

%paper roadmap 
Our paper is structured as follows. We outline related work around privacy policy research, risks associated with data aggregation, DBDPs, and data valuation in Section \ref{section:background}. We investigate the PII data types lexicon on the temporal privacy policy dataset measured by its summary statistics, stability, complexity, and sensitivity over time. We also report the results of our mesoscale analysis on the structure and dynamics of the policies over time in Section \ref{section:results}. We discuss our findings, their broader implications, and future work in Section \ref{section:discussion}. Finally, We outline a large temporal dataset of privacy policies and a lexicon of PII relevant terms and explore methods used in Appendix \ref{appendix:I}, Appendix \ref{appendix:II}, Appendix \ref{appendix:III}, and Appendix \ref{appendix:IV}.

\section{Background}
 \label{section:background}

\subsection{Research on Data Brokers} Otto et al. \cite{otto2007choicepoint} conducted a critical case study of ChoicePoint, a data broker well known for a massive data breach and consumer information mishandling. In 2005, ChoicePoint disclosed that it had sold the personal records of thousands of individuals to identity thieves, resulting in hundreds of fraud and identity theft cases. In turn, this scandal shed light on ChoicePoint's practices and gave the researchers access to data about the firm's activity that would not otherwise be available.

In the case study, the authors track the flow of data collected, bought, and sold by ChoicePoint. Their analysis of how the data flows from one place to another highlights the extensive breadth of data sources collected, purchased, and sold by the broker. It also highlights that the data types sold were highly sensitive (including D.N.A. sequences and social security numbers). These data were sold, without restriction, to a wide variety of buyers, which ranged from individuals to large agencies. The only reason the authors have such a detailed account of the data flow for this case study is that ChoicePoint had several large data breach events. Brokers who have not faced large events such as bankruptcy or security breaches are not required to disclose this information. Roderick and Crain \cite{crain2018limits,roderick2014discipline} argue that this lack of transparency and an asymmetry in power takes away consumers' ability to protect their data. More transparency would open up additional analysis of this industry. 

\subsection{Research on Risk and Harms of Combining Data Types} The harms associated with collecting data increase as different types of data are aggregated together. The harms become more than the sum of their constituent parts. For example, a dataset of names may not be considered sensitive. Once combined with associated social security numbers, the data becomes much more sensitive than each data point on its own. Further, if combined with a personal address, this information collection becomes more sensitive because it can be used to assume a person's digital identity. With the collection of even more information, such as browsing history, this aggregate of information can be used to build a digital dossier \cite{solove2004digital} on the data subject, which can be used to make inferences about them, predictions about their future behavior, and manipulate their future behavior with targeted advertising. Datasets become more sensitive and pose more privacy risks to a subject as more data types are combined.

It is well studied that aggregation of PII and associated data types increase the data privacy risk to the data subject. \cite{wagner2018privacy} Privacy risk pertains particularly to the potential or actual harm to individual data subjects. According to Wagner et al., \cite{wagner2018privacy} the impact of privacy risk can be broken down into composite categories: scale, sensitivity, expectation, and harm.  Scale can be quantified by the number of individuals implicated. Sensitivity can be quantified by the number of data types involved, entropy, and the average privacy setting. The expectation of risk can be quantified as deviation from the expectation of how the data will be handled. Finally, harm can be quantified as damages awarded or perceived harm.

In this project, we will primarily focus on the impact of sensitivity measured by the number of different PII data types collected together over time by DBDPs. As stated above, aggregations of different data types display a higher level of risk than the sum of the component data type risks in isolation. In future work, we will measure harm through tort damages as defined by Wagner et al. 

\subsection{Analysis of Privacy Policies} The current over-reliance on data subjects to take on the burden of reading privacy policies and consent online has led to a negative externality of less legitimate consent due to consent fatigue. \cite{schermer2014crisis} McDonald et al. \cite{mcdonald2008cost} estimate that if every individual were to fully read every privacy policy they encountered online in the United States during 2008, the national opportunity cost would be 781 billion dollars.

Several researchers have built automated methods to help data subjects parse through the legal language of privacy policies in shorter time periods. \cite{srinath2020privacy, cranor2003p3p} We believe these are indeed very important tools. Still, we also recognize that sometimes the devil is in the details regarding consent agreements. A balance between easily understood and readable versions \cite{graber2002reading, pollach2007s} of privacy policies and thorough disclosure of what information is being collected, used, and shared is also needed. 

Other studies have investigated the text of privacy policies to understand how they have adapted to legal frameworks over time \cite{linden2018privacy,zaeem2020effect, stallings2020handling, amos2021privacy} and how well they align with consumer values. \cite{earp2005examining} In 2021, Amos et al. \cite{amos2021privacy} crawled over one million privacy policies on the Way Back Machine from the past 20 years to bring transparency to these issues. This dataset is made publicly available by request. They also created an automated trend detection tool to identify terms and concepts that show shifts in the language of privacy policies. Linden et al. \cite{linden2018privacy} also conducted a text analysis on privacy policies to examine how these policies have changed since the introduction of the EU General Data Protection Regulation (GDPR). In 2020, Srinath et al. created the web crawler \textit{Privateer}, \cite{srinath2020privacy} which made a large-scale corpus of web privacy policies.

\section{Results}
 \label{section:results}

Our results can be broken up into three mesoscale categories:  

\textbf{Result \#1. Word Level:} In result  \#1, we highlight the stability of the privacy policy PII data type terms over time as represented by frequency trends. Several sensitive data types related to location, behavior, and internet activity appear to rise over time. Stable words predominantly describe customer information. Falling words may result from other, more specific words taking over. For example, ``coordinates'' appears to be falling in frequency over time, but other geolocation words are taking over the word space. For the most part, new words appear to reflect biometric information that may be tied to new technology or methods for inference, such as ``faceprint'' and ``voiceprint.''

\textbf{Result \#2. Topic Level:} In result \#2, we explore the complexity of the privacy policies as represented by the compression factor of privacy policies compared by year. We see a steady decrease in complexity over time from 2000-2019. In our analysis of topic prevalence over time, the usage of topics is relatively stable over time, which speaks to the rigid and established language used in privacy policies. Moreover, we see trends related to new technologies and legislation as privacy policies are added and amended over time.

\textbf{Result \#3. Network Level:} In result \#3, we investigate the level of sensitivity and risk (\# of PII data types collected together) in the privacy policies as represented by the density of the word co--occurrence network graph. We also consider the network's density alongside modularity because there are networks of different sizes across time. Modularity allows for a quantitative comparison of community structures via the number of classes across networks of different sizes. We see that network density rises over time.

% \textbf{Result \#1. Word Level:}
% \textbf{Result \#2. Topic Level:} 
% \textbf{Result \#3. Network Level:}
% Q1: What PII words, topics, and co--occurrence network structures appear in privacy policies over time?
% Q2: What PII data types are collected from consumers, and how stable is the representation of those PII data types
% over time?
% Q3: How complex or regular are privacy policies over time?
% Q4: What PII data types are collected concurrently from consumers, and how has the level of sensitivity (and potential
% risk to data subjects) changed over time?

We address our high-level research questions as follows. Question 1: What PII words, topics, and co--occurrence network structures appear in privacy policies over time? Will be addressed by results \#1-3. Question 2: What PII data types are collected from consumers, and how stable is the representation of those PII data types over time? Will be addressed in result \#1. Question 3: How complex or regular are privacy policies over time? Will be addressed in result \#3. Finally, question 4: What PII data types are collected concurrently from consumers, and how has the level of sensitivity (and potential risk to data subjects) changed over time? will also be addressed in result \#3. 

\subsection{Descriptive Statistics of Privacy Policy Data}
\label{section:descriptivestats}

The privacy policy corpus is provided by Amos et al. \cite{amos2021privacy} in SQLite format and weighs in at 48.24GB. It includes over 1 million snapshots of privacy policies from the Internet Archive from 1997-2019. The data includes many fields, but for our purposes, we use the following columns: the full text of the privacy policy snapshot in markdown, the year it was collected, the URL, and the category of the institution (as classified by Webshrinker’s API) for the policy. The necessary information was extracted from the corpus using SQL queries and converted to a comma-separated values file format using the Pandas library. On average, a website has 8.4 privacy policy snapshots ($M=6$). $79.4\%$ of snapshots are from 2010 or later. An overview of summary statistics about the privacy policy dataset can be found in Table \ref{tab:corpusdata} in Appendix \ref{appendix:I}.

To explore the dataset for our research questions, we cleaned and segmented the data in several ways: (1) We filter all sentences that include negation (negation words can be found in Appendix \ref{appendix:III}); (2) Used NLTK for Tokenization and removal of punctuation (Regexp \texttt{\textbackslash w}); (3) Made all text in the privacy policies lowercase; (3) Split the dataset by year; (4) For some analysis, we filter all of the corpora to extract just the terms in our PII Lexicon. PII lexicon words can be found in Appendix \ref{appendix:II}. We decided to keep non-sentence text, like headings or tables, and treat it the same as other parts of the text, as they could potentially use slightly different wording and thus help us match on specific terms.

\subsection{PII Data Types Lexicon}
\label{section:piidatatypes}

For our project, we manually curated a lexicon of 287 terms that are associated with personally identifiable information (PII) as defined by legislative definitions sections from nine U.S. state-level laws concerning consumer data privacy. These laws include: 

\begin{enumerate}
    \item California Consumer Privacy Act (CCPA) of 2018 (Cal. Civ. Code §§ 1798.100 et seq.)
    \item California Consumer Privacy Rights Act (CPRA) of 2020 (Proposition 24 A.B. 1490)
    \item Colorado Privacy Act of 2021 (Colo. Rev. Stat. § 6-1-1301 et seq)
    \item Connecticut Act Concerning Personal Data Privacy and Privacy Online Monitoring 2022 (Senate Bill No. 6 File No. 238 Cal. No. 189) 
    \item Utah Consumer Privacy Act 2022 (S.B. 227) 
    \item Virginia Consumer Data Protection Act 2021 (2021 H.B. 2307/2021 S.B. 1392) 
    \item California Data Broker Registration Civil Code 2019 (Cal. Civ. Code §§ 1798.99.80 et seq)
    \item Nevada internet privacy laws 2021 and 2017 (NRS § 603A.300 and  2021 S.B. 260)
    \item Vermont Protection of Personal Information Act 2017 (9 V.S.A § 2446-2447). 
\end{enumerate}
To create our lexicon, we manually extract defined terms by reading legislative documents and collecting words defined as personally identifiable information (PII) in all legislative documents' definitions sections. The full lexicon of PII data type terms can be found in Appendix \ref{appendix:II}. PII lexicon summary statistics can be found in Appendix \ref{appendix:I}. An overview of descriptive statistics on the filtered PII data types corpus can be seen in Appendix \ref{appendix:I} Table \ref{tab:piicorpusdata}. Note that the number of unique words present in each year shows the number of unique PII data types used in the corpus for that year which increases steadily over time.

%  \begin{figure} 
%     \centering
%   \includegraphics[width=0.6\columnwidth]{Figs/no_pii_data_types.pdf}
%   \caption{Number of unique PII data type terms present in the PII filtered corpus by year}
%   \label{fig:piidatatypes}
% \end{figure}

\subsection{Word level results: Measures of turbulence}

In this section, we look at the frequency distribution of our PII data types lexicon as a time series to investigate what PII data types demonstrate rapid increases in frequency, decreases in frequency, stability in frequency, and introduction of new words over time. We will also look at groups of related words to demonstrate how the PII data types can be used to query related groups of words to investigate their frequency distributions in the context of more specific legislative events. 

\begin{figure} 
  \centering
  \includegraphics[width=0.45\columnwidth]{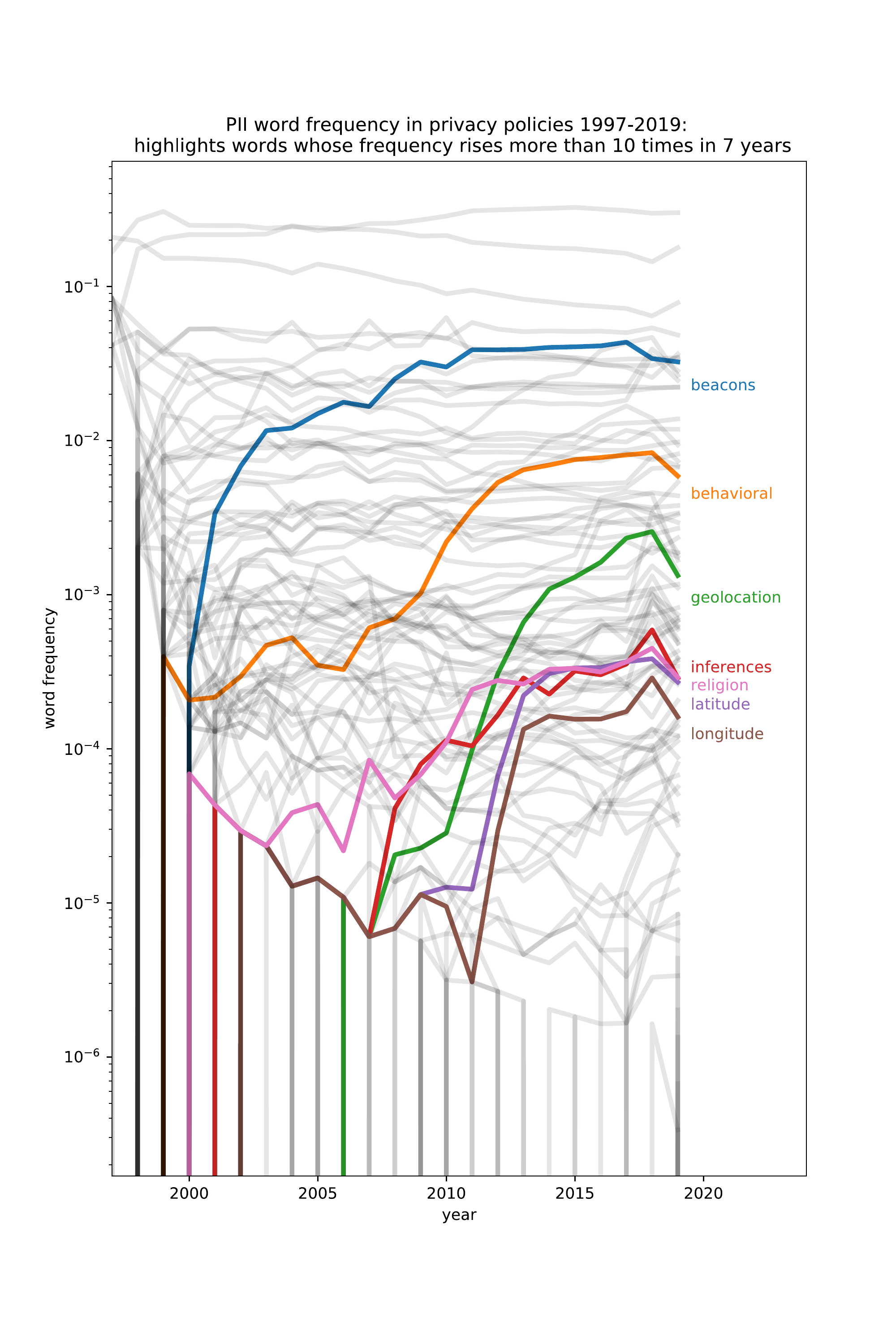}
  \includegraphics[width=0.45\columnwidth]{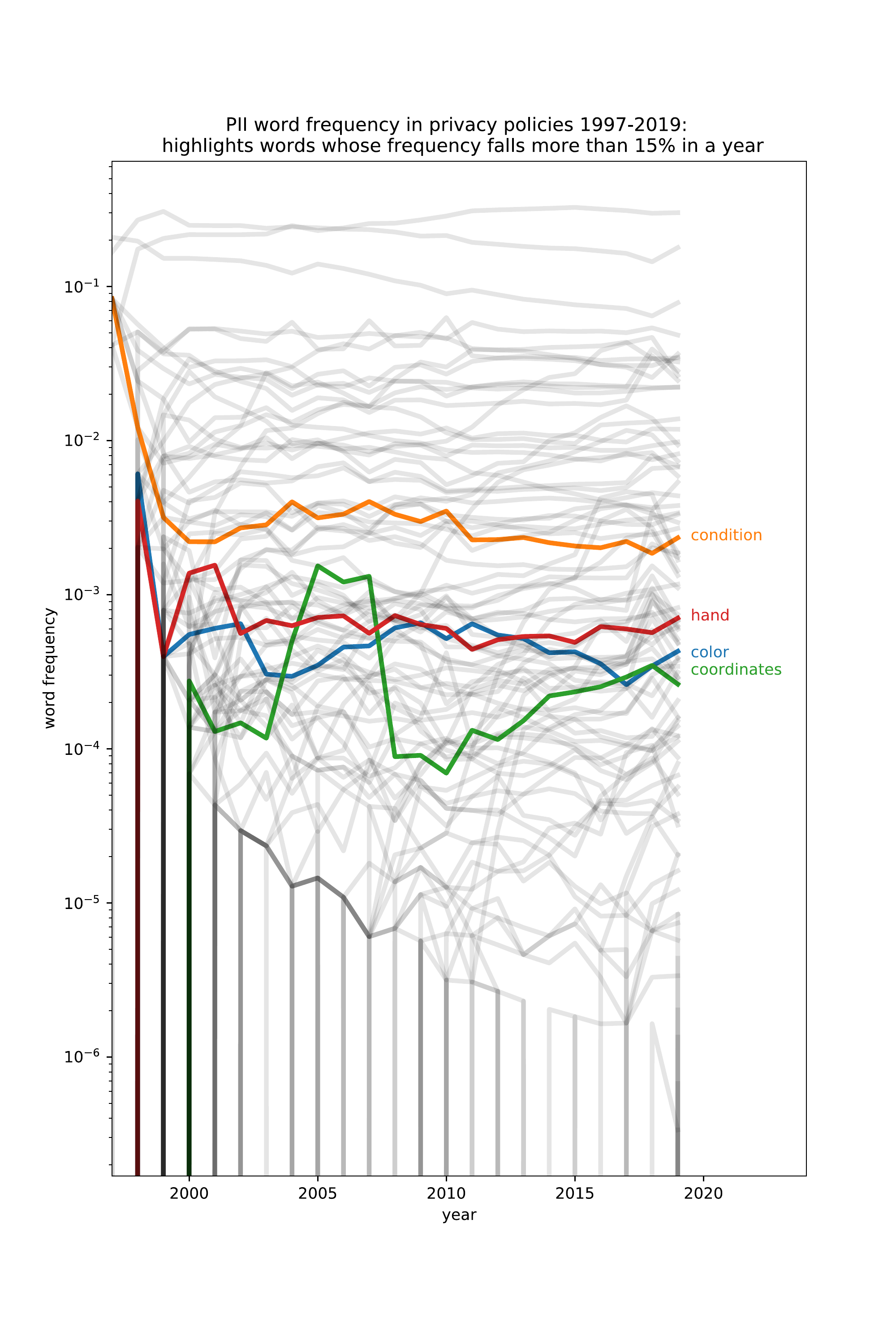}
  \caption{a.) Frequency Distribution of words that rise more than ten times in a 7-year period, b.) Frequency Distribution of words that fall more than 15\% in a year}
  \label{fig:rise_fall}
\end{figure}

\textbf{Rise:} Rising looks at PII data types whose frequency increases quickly over time. We define rising words as words that rise in frequency by more than ten times in a 7-year period, as seen in Figure \ref{fig:rise_fall}a. Our results show that PII terms such as ``beacons,'' ``behavioral,'' ``geolocation,'' ``inferences,'' ``religion,'' ``latitude,'' ``longitude'' rise quite dramatically in the dataset over time. It is worth noting that these terms represent rather sensitive types of information related to geolocation data, browsing behavior, and inferences. It is noteworthy that these data types represent information collection methods that are indirect, meaning it is not information directly given to the DBDPs but is gathered from monitoring the data subject's behavior and digital traces over time.  

% \todo{add more narrative on the importance of these words}. 

\textbf{Fall:} Falling looks at PII data types whose frequency drops quickly over time. Falling words are defined as words that fall in frequency more than 15\% in a year, as seen in Figure \ref{fig:rise_fall}b. Our results show that terms such as ``condition,'' ``hand,'' ``coordinates,'' and ``color'' fall quite dramatically in the dataset over time. The decreased use of words like ``coordinates'' may be explained by the rising use of other geolocation words taking their place in the word space as seen in the rising words results. 

\begin{figure} 
    \centering
  \includegraphics[width=0.45\columnwidth]{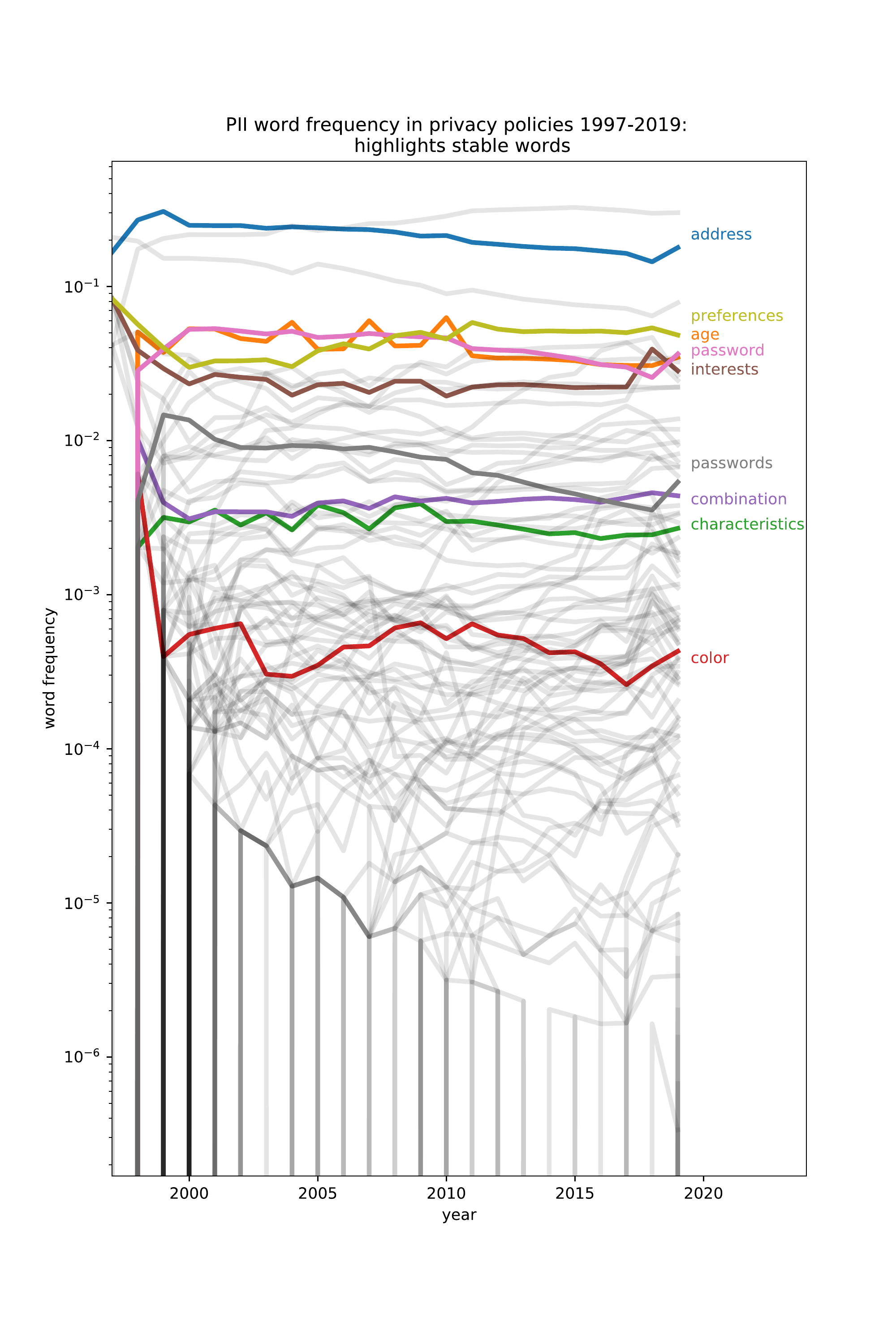}
  \includegraphics[width=0.45\columnwidth]{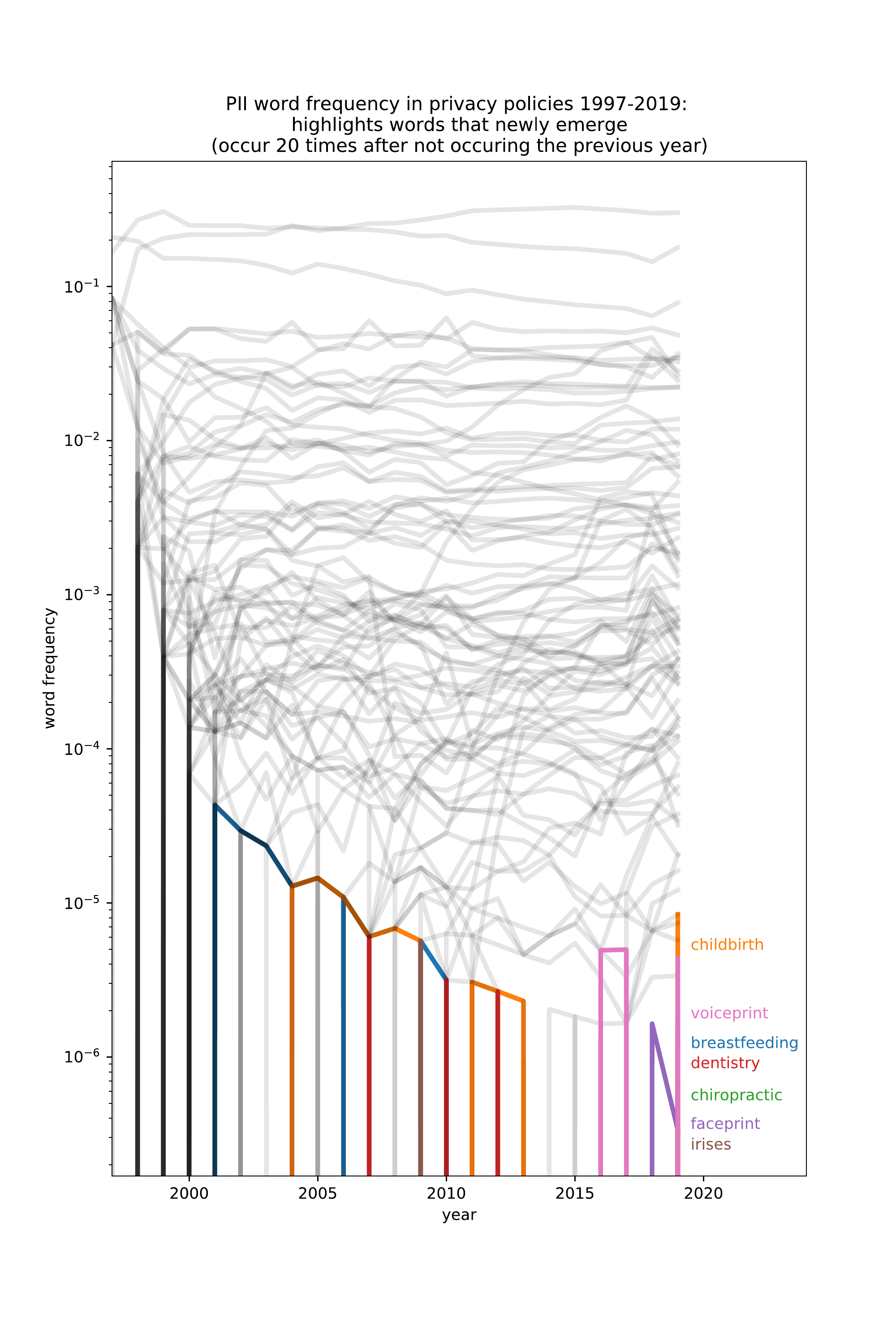}

  \caption{a.) Frequency distribution of stable words (change less than 2\% in frequency over a 20-year period), b.) Frequency distribution of words that emerge (occur 20 times after not occurring the previous year)}
  \label{fig:stable_emerge}
\end{figure}

\textbf{Stability:} Stability explores PII data types whose frequency distributions remain at similar frequencies over time with little change. Here stability is defined by the frequency distribution of words that change less than 2\% in frequency over a 20-year period, as seen in Figure \ref{fig:stable_emerge}. Our results show that PII terms such as ``address,'' ``preferences,'' ``age,'' ``password,'' ``interests,'' ``passwords,'' ``combination,'' ``characteristics,'' and ``color'' remain steadily present in the privacy policies over time. These words generally represent consumer identifiers and characteristics which would be expected for DBDPs to collect while conducting regular business with data subjects. 

\textbf{Emergence:} Emergence looks at new PII data types that appear in privacy policies over time. We define emergent words as words that occur at least 20 times in one year after not occurring the previous year, as seen in Figure \ref{fig:stable_emerge}. Our results show that PII-relevant terms such as ``childbirth,'' ``voiceprint,'' ``breastfeeding,'' ``dentistry,'' ``chiropractic,'' ``faceprint,'' ``irises'' newly emerge in the dataset primarily in the years 2016-2019. The words we see emerging appear to represent new biometric technologies and words related to health.

\textbf{Frequency distribution of health-related words:} We can use the frequency distribution of PII terms to explore groups of related words. We explore groups of words related to health and inference as an example to showcase how the time series of these frequency distributions may be used to better understand sub-topics of PII data types. Our first example explored PII data types related to health, and we compared the frequency distribution to potentially correlated legislative events. Our results show that a higher frequency distribution of words related to health and medicine may correlate with events related to the Health Insurance Portability and Accountability Act (HIPAA). HIPAA restricts the healthcare industry from disclosing protected health PII without patient consent. 

\begin{figure} 
  \centering
  \includegraphics[width=0.5\columnwidth]{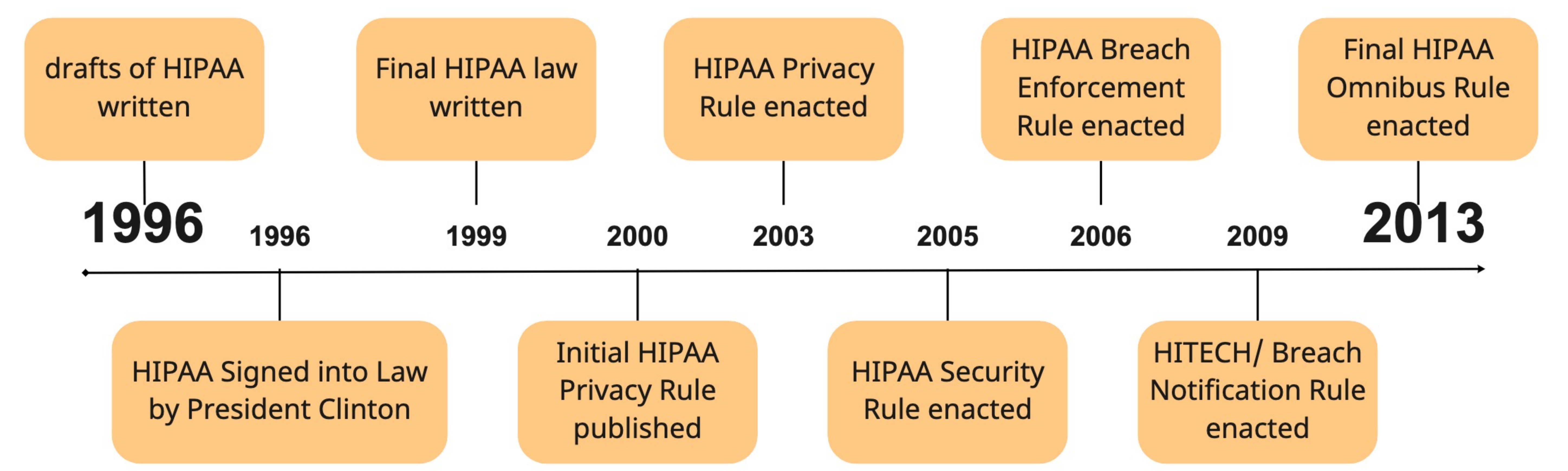}
  \caption{HIPAA Timeline and History}
  \label{fig:hipaatimeline}
\end{figure}

\begin{figure} 
  \centering
  \includegraphics[width=0.9\columnwidth]{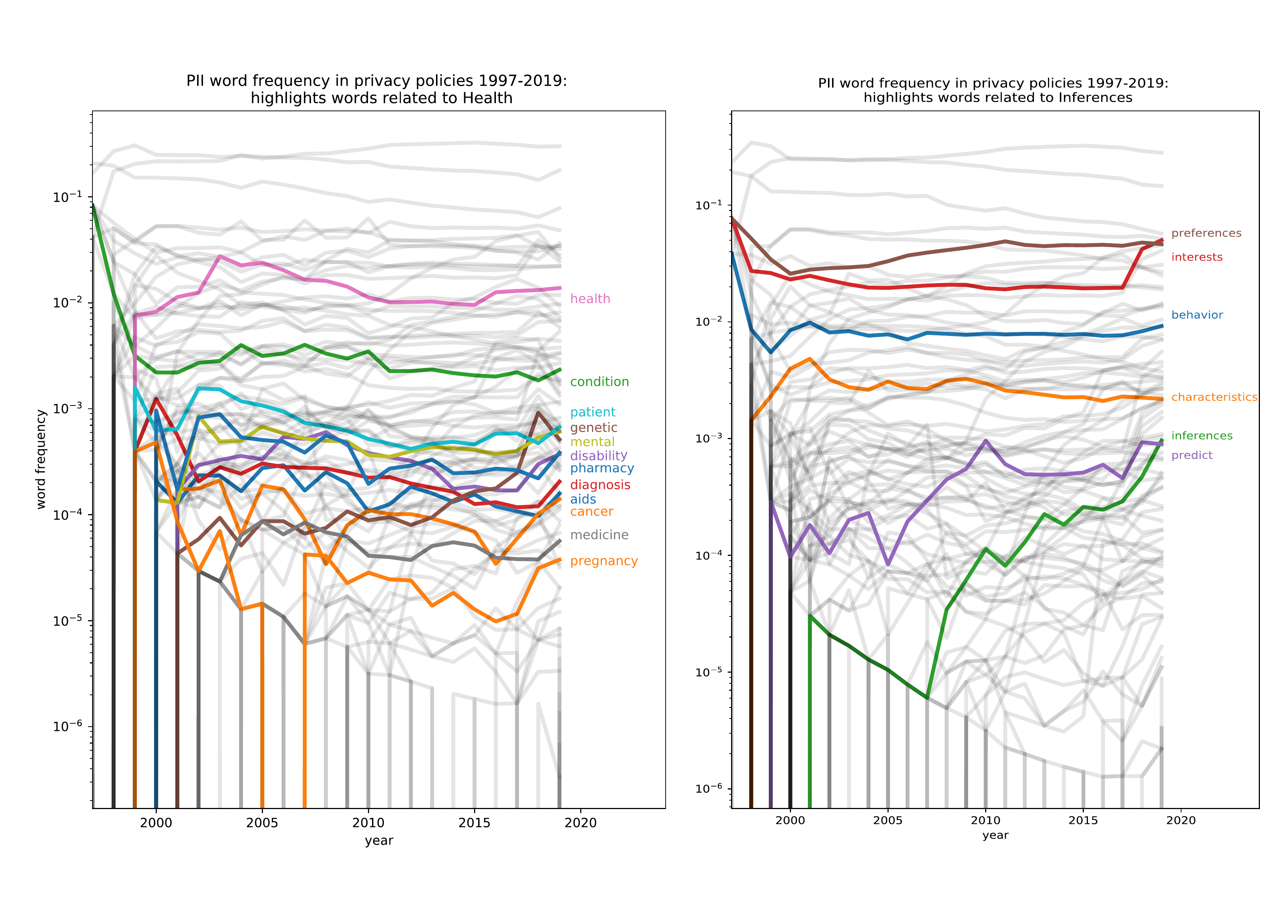}

  \caption{a.) Frequency distribution of words related to health, b.) Frequency distribution of words related to insights}
  \label{fig:healthinference_words}
\end{figure}

HIPAA has a long and complicated history which begins with its introduction in 1996 and final enactment in 2013. A timeline of notable HIPAA events can be found in Figure \ref{fig:hipaatimeline}. Comparing our HIPAA timeline with frequency distributions of health-related words, as seen in Figure \ref{fig:healthinference_words}a, we do see spikes in the use of health-related words during periods when HIPAA regulations were being rolled out (note that there will be a natural delay between legislation being enacted and privacy policies reacting to them). We can see words like ``health,'' ``patient,'' and ``pharmacy'' start to spike around 2002-2004. A notable exception in the frequency distribution behavior of health words is the word ``genetic,'' which rises steadily from 2001-2019.

% \begin{itemize}
%     \item 1996 - drafts of HIPAA written 
%     \item 1996 – HIPAA Signed into Law by President Clinton.
%     \item 1999 - Final HIPAA law written
%     \item 2000 - Initial HIPAA Privacy Rule published
%     \item 2003 – HIPAA Privacy Rule enacted
%     \item 2005 – HIPAA Security Rule enacted
%     \item 2006 – HIPAA Breach Enforcement Rule enacted
%     \item 2009 – HITECH and the Breach Notification Rule enacted
%     \item 2013 – Final HIPAA Omnibus Rule enacted
% \end{itemize}

% \begin{figure} 
%   \centering
%   \includegraphics[width=\columnwidth]{Figs/2001_2000_shifterator.png}
%   \caption{Shifterator proportional rank shift divergence}
%   \label{fig:shifterator}
% \end{figure}

% \textbf{Shifts in location-related words:}

% Our second example of frequency distributions of word groups explored location PII data types. In the frequency distributions, location words emerged in 2001 and 2002 but only rose to the highest frequency in 2017-2019. The frequency distribution of location-related words can be found in Figure \ref{fig:location_words}. In this figure, the PII data type ``location'' from its emergence in 1998 has risen continuously through 2019. The term ``locate'' peaks in 2004. The PII term ``coordinate''' peaked drastically between 2003-2005. Notably, the term ``geolocation'' from its emergence in 2007, has risen drastically through 2019. The word ``latitude'' and ``longitude'' (often used together), from their emergence in 2002, dropped until 2008 and now continue to rise through 2019, notably, the usage of the two words maps quite closely to one another. Interestingly, the word ``radius'' emerged in 2004 and peaked in 2006, but it appears to be a more turbulent term. 

% \begin{figure} 
%   \centering
%   \includegraphics[width=0.5\columnwidth]{Figs/location_pii_allyears.pdf}
%   \caption{Frequency distribution of words related to location}
%   \label{fig:location_words}
% \end{figure}

\textbf{Frequency distribution of insights-related words:} As we mentioned earlier, Insights are ways in which DBDPs can collect PII data and sell the insights gleaned from the PII data as a way to gain from the data collection financially but avoid being classified as a data broker. In Figure \ref{fig:healthinference_words}b, we look at words that are related to inferences and insights. Most insight words are used quite frequently compared to other words and remain fairly stable with two notable exceptions, ``predict'' is quite peaky and spikes to its highest point in 2010 but has been on the rise since its emergence in 1999. The term ``inferences'' emerged in 2001 and  began to rise rapidly starting in 2007.

\subsection{Topic level results: Measures of complexity}
\label{section:complexityresults}

\begin{figure}
  \centering
  \includegraphics[width=0.7\columnwidth]{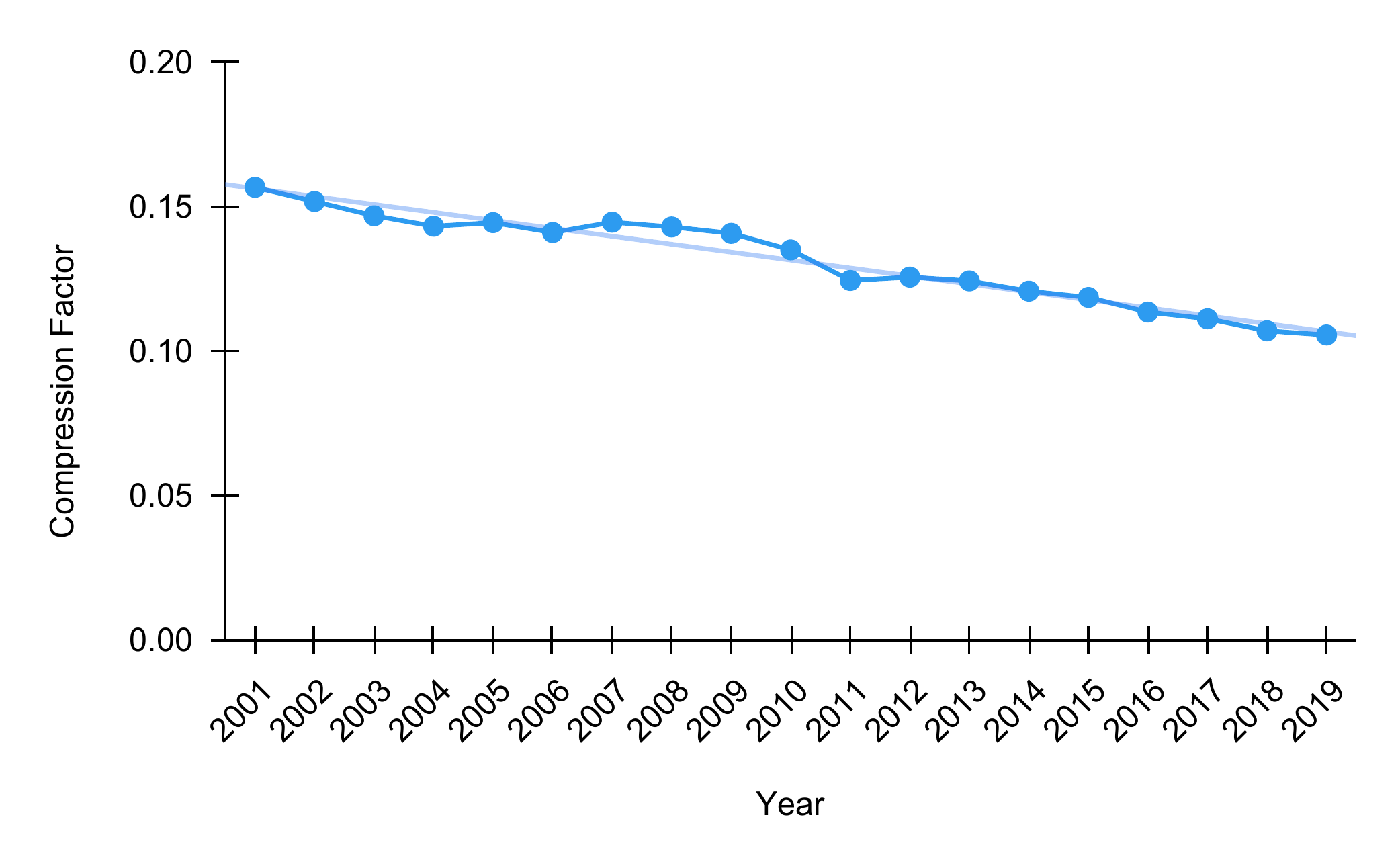}
  \caption{This figure represents the Privacy policy policies 1997-2019 complexity measured by year via a compression ratio. The blue line indicates the compression factor from the full text of each year, and the pink line indicates the compression factor from a random sample of 2,800 policies selected for each year. Smaller compression factors mean the model can compress the corpus more. To compute the compression ratio, we get the Minimum Description length (MDL) and convert the output from nats into bits (MDL bits = MDL nats$/\textnormal{log}_{e}2$). We then get the size of the original file in bits which we calculate as the total number of words in the corpus multiplied by the log of the number of unique words in the corpus. \cite{hebert2022network, gerlach2018network} To compute the compression ratio, we divide the MDL in bits by the original file size in bits. We then compare the compression ratio in bits over time for the privacy policies as extracted from year-by-year corpora using the hSBM topic model. \cite{gerlach2018network}}
  \label{fig:complexityperyear}
\end{figure}

\textbf{Topic Complexity:} In result \#2, we explore the complexity of the privacy policies as represented by the compression factor as seen in Figure \ref{fig:complexityperyear} (full details can be seen in Appendix \ref{appendix:I} Table \ref{tab:corpusdata}. We see a trend of the compression factor \cite{hebert2022network} going down over time which can be interpreted as the privacy policies containing more regularity \cite{van1994measuring, friedrich2021complexity} (meaning the policies look more similar to one another) and therefore, more compress-able over time. The compression factor in our model signals the complexity of the privacy policies for that year (compression ratio = bits of the Minimum Description Length (MDL) divided by bits of a non-compressed description). We see a fairly steady decrease in the complexity of the policies over time, which may be related to new regulatory pressures which require policies to disclose certain information in more precise ways (e.g., as CCPA, GDPR) to comply. In the near future work, we would like to test if we can use the compression factor as a means to detect individual privacy policies or clusters of privacy policies which exhibit irregularities as a possible means of detecting non-compliance. 

\subsection{Topic level results: Topic prevalence over time}
\label{section:topicsrprevresults}

\begin{figure*} 
  \centering
  \includegraphics[width=\textwidth]{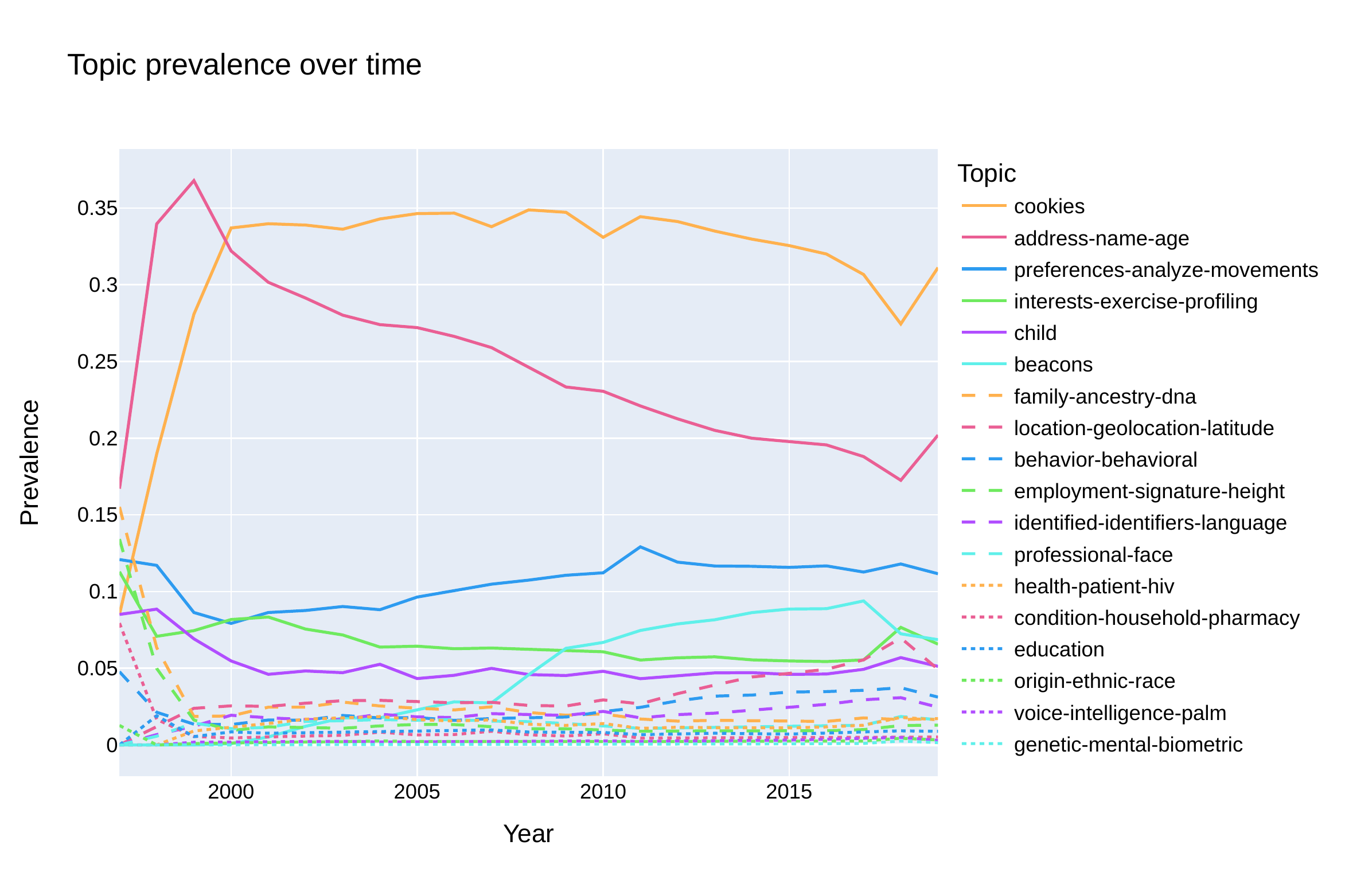}
  \caption{Topic prevalence 1997-2019. Topics extracted from the full corpus after negation filtering through hSBM topic modeling.}
  \label{fig:topicsprev}
\end{figure*}

In our time-series analysis of topic prevalence, we find several trends that confirm our previous findings (seen in Figure \ref{fig:topicsprev}). We find ``cookies'' to dominate the topic landscape throughout the entire time frame. Manually entered PII attributes like address, name, and age show high prevalence but dwindle over time as newer tracking mechanisms emerge and enable the collection of more valuable data. Behavioral data like preferences and movement are the third most prevalent topic overall and rise slightly over time.

Interestingly we see the topic ``beacons'' rise rapidly from 2007 through 2009, which correlates with the introduction of Facebook Beacon in 2007 \cite{jamal2013mining} and the shutdown of Facebook Beacon after a class action lawsuit in 2009 that claimed the use of Facebook beacons violated user privacy. Notably, the topic prevalence (and word frequency distribution) of beacons remained fairly steady from 2011-2017, with dips occurring from 2018-2019. This points to updates to privacy policies being more often additive than subtractive - overly broad policies do not seem to cause serious problems to the organizations.

In addition, we find ``location-geolocation-latitude'' rising significantly in prevalence from 2011 on, with slope increasing over time. This coincides with a steep rise in the availability of smartphones after the iPhone, and the first Android models were introduced in 2008. The Android Open Source Project opened access to the smartphone market to smaller companies in 2010. Smartphones drove mobile usage of many digital services and contained GPS locators and thus making much richer location information available than stationary or laptop computers did previously.

We find usage of the most prevalent topics, except ``address-name-age'', stable over time, which is explained by the narrow purpose of an established, formal language used in privacy policies. The top five topics prevalent over time are as seen in Figure \ref{fig:topicsprev}:  cookies, address-name-age, preferences-analyze-movement, interests-exercise-profiling, and child. The top two topics dominate the whole time frame. However, they decrease in prevalence over time, which points to privacy policies being used for many more different purposes in 2019 than 20 years earlier. A list of vector representations of all analyzed topics can be found in Appendix \ref{appendix:IV}.

\subsection{Network level results: Measures of sensitivity}

In result \#3, we investigate the level of sensitivity and risk (\# of PII data types collected together) in the privacy policies as represented by the density of the word co--occurrence network graph in relation to modularity and the number of classes. We choose these two network measures to represent the sensitivity of the co--occurrence in the network because they represent the proportion that words are presented in privacy policies together and can be seen as a representation of the number of PII data types collected together. 

\textbf{Network density} is defined as the proportion of the number of edges in a network compared to the number of potential edges between all pairs of nodes in the graph. Both the measure of network density and modularity will allow us to examine a fuller picture of the graph density qualified by the modularity density, which accounts for the network's density in relation to the size of the network. 

We see that density in the network rises over time. That density and modularity show spikes during the critical time when new privacy legislation is introduced. As we use these network density measures as a representation of sensitivity, we can say that the co--occurrence of PII data type words does appear to be increasing. However, it is unclear whether this is a measure of increased concurrent PII data type collection activity or perhaps just a signal of compliance with new legislative mandates for more precise PII data type disclosure. It will be interesting, looking forward, to see whether this trend continues. 

\begin{figure*}
  \centering
  \includegraphics[width=0.49\columnwidth]{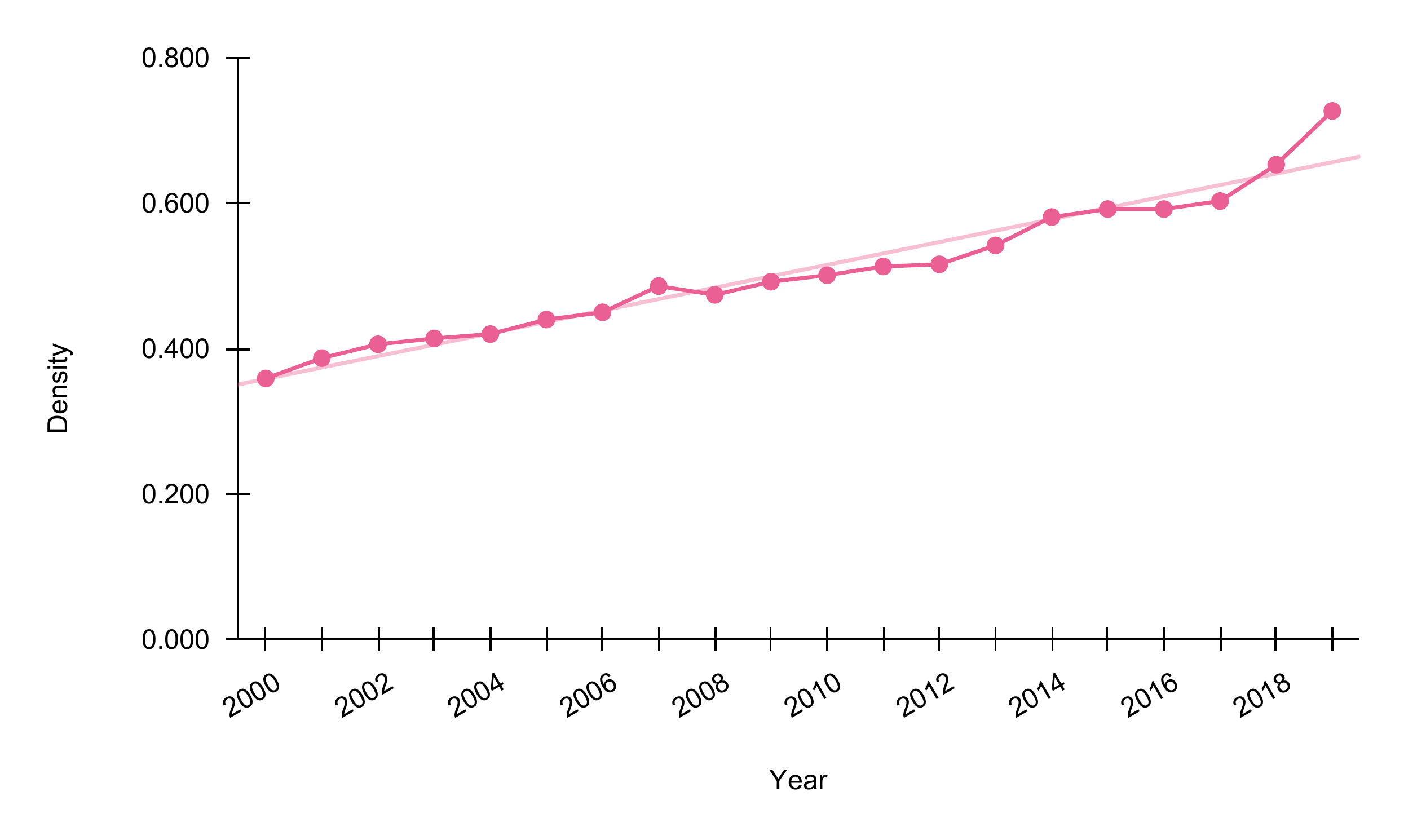}
  \includegraphics[width=0.49\columnwidth]{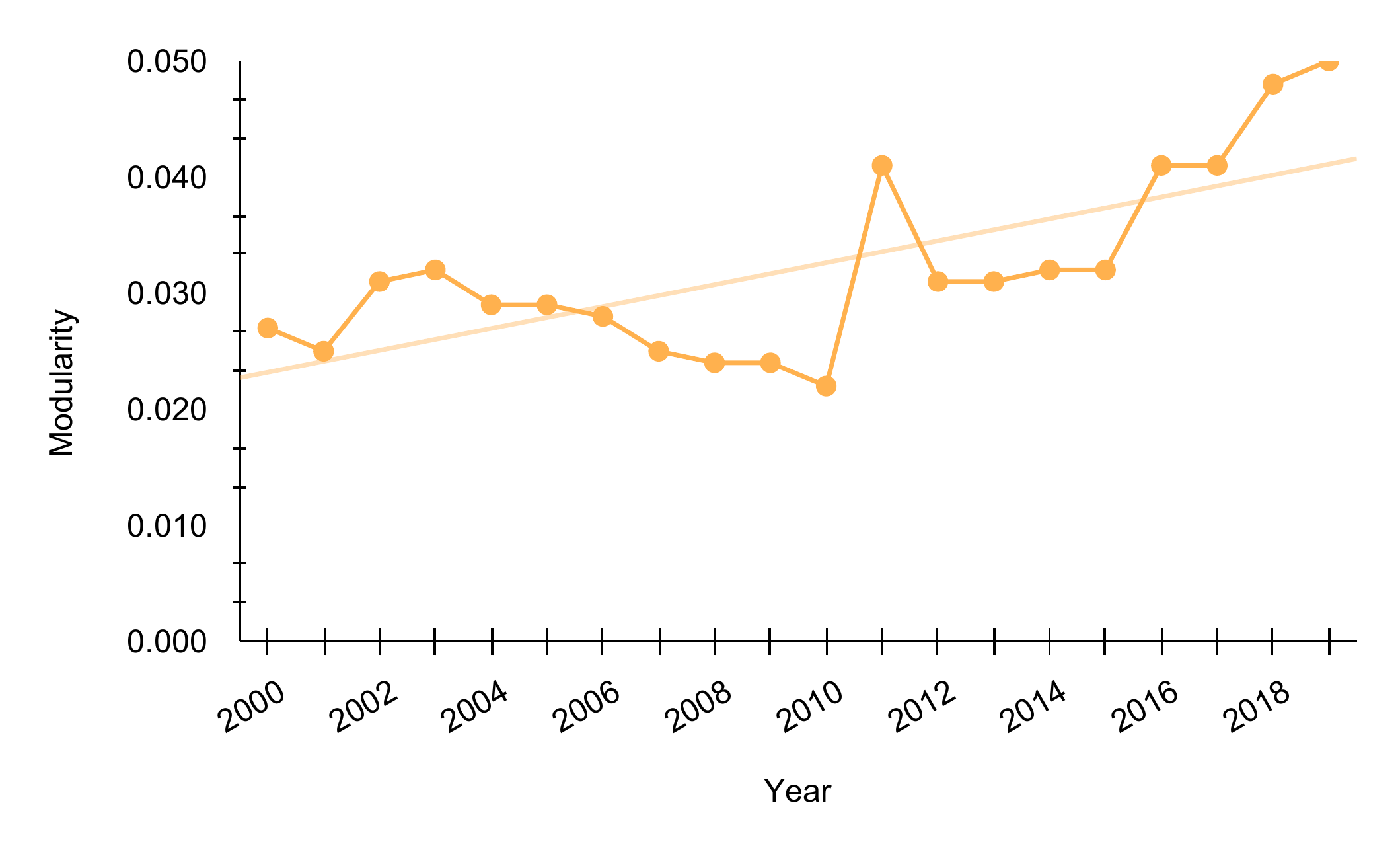}
  \caption{a.) 2000-2019 co--occurrence network of PII terms network density by year. b.) 2000-2019 co--occurrence network of PII terms network modularity by year.}
  \label{fig:density-modularity}
\end{figure*}

In Figure \ref{fig:density-modularity}a, we can see between 2000-2019,  there is a fairly steady increase in network density (we exclude the years 1997-1999 because their networks are too small). We can also see that there is an increase in density from 2013-2019. This makes sense in light of our timeline of legislative policy changes in the US and EU (as seen in fig. \ref{fig:changes-timeline}). During 2012-2019, we saw wide-sweeping data privacy laws come onto the scene, which requires that DBDPs begin to disclose more information (CCPA in 2018, for example, requires they list the data types they collect) in their privacy policies.

\textbf{Network modularity} is a scalar value that ranges from $-1$ to $1$ and measures the density of links within communities compared to that of links between the communities. \cite{blondel2008fast} Modularity measures the strength of the division of a network into classes. The modularity coefficient is usually computed by finding an optimal partition of the network into classes.

A high modularity score means the network has dense connections between nodes in the same class but sparse connections between nodes in different classes. In Figure \ref{fig:density-modularity}b, we can see a rise in modularity between 2002-2005 (which coincides with the E-government act of 2002, the State Data Breach Notification Laws of 2003, and the California Online Privacy Protection Act of 2004). Modularity then went down again between 2006-2010 (which is a period with no new broad sweeping privacy laws), and then in 2011, there was a very large spike. From 2012-2019, we see a steady yearly increase in modularity (which coincides with several very large sweeping privacy laws such as GDPR and CCPA). Additional network measures can be seen in Appendix \ref{appendix:I} Table \ref{tab:descdfdc2}.

% \begin{figure} 
%     \centering
%   \includegraphics[width=0.5\columnwidth]{Figs/.pdf}
%   \caption{2015 co--occurrence network. Backbone threshold at 0.1. All nodes with a degree less than one are filtered from the network (words that do not co--occur with other words). Networks are partitioned by modularity, signified by the nodes' color. The node size ranges from 1 to 40 based on the weighted degree.}
%   \label{fig:coonet_2015}
% \end{figure}

We also analyze the degree and strength distributions of the co--occurrence graph for 2019 (our largest year), based on negation-filtered data, and compare them to the distributions found by Fudolig et al. in. \cite{fudolig2022sentiment} They find power laws in degree and strength distributions in an analysis of word co--occurrence in tweets.

\begin{figure}
  \centering  
  \includegraphics[width=0.45\textwidth]{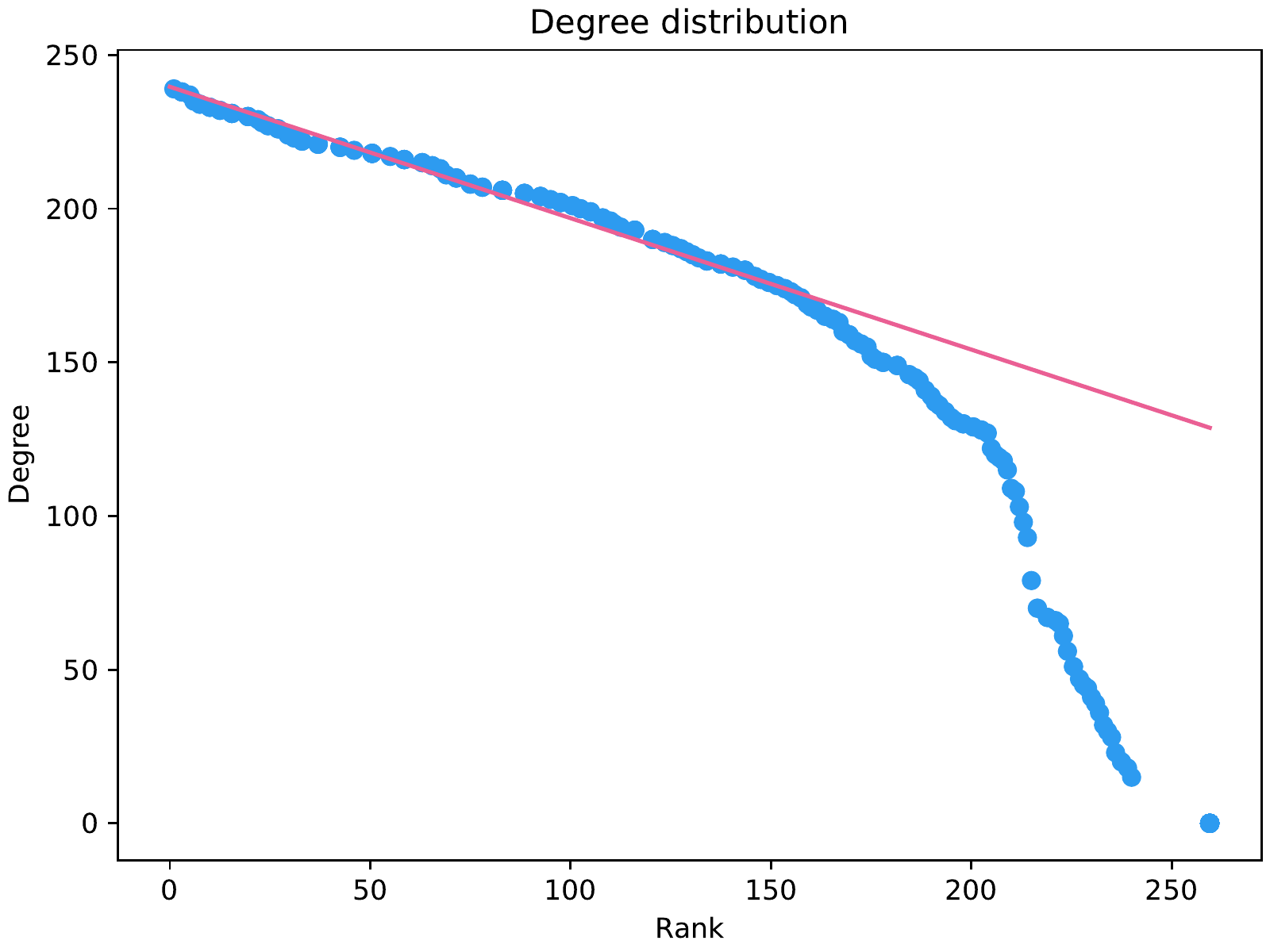}
  \includegraphics[width=0.45\textwidth]{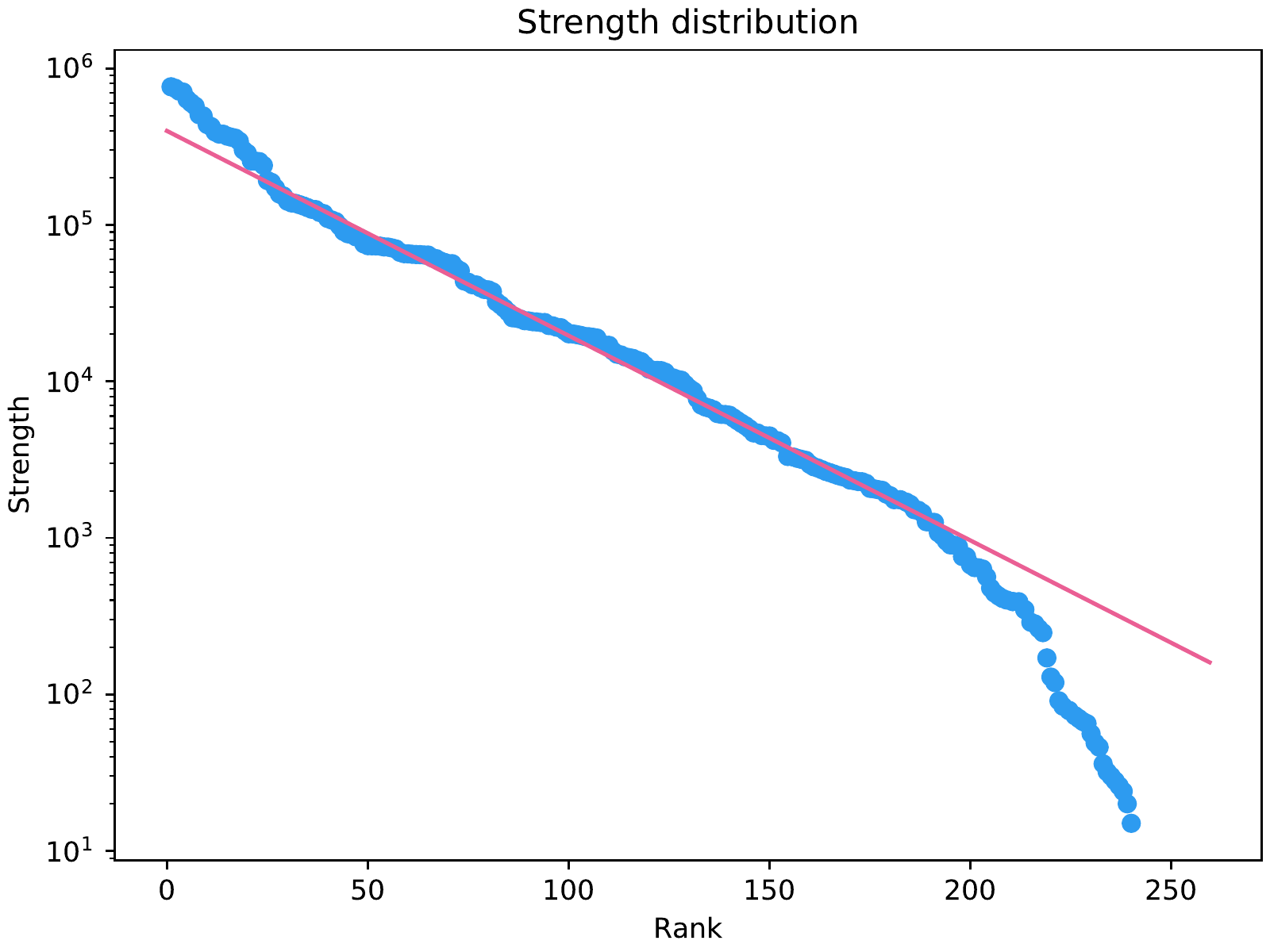}
  \caption{a.) The linear plot is a Zipf-ranked degree distribution of the 2019 co--occurrence network of PII terms. b.) Zipf-ranked strength distribution of the 2019 co--occurrence network of PII terms, logarithmic $y$ axis}
  \label{fig:co--occur2019-degree-dist-strength-dist}
\end{figure}

As shown in Figure \ref{fig:co--occur2019-degree-dist-strength-dist}, we do not find anything close to a power law in either. We find an approximately linear degree distribution and an approximately exponential strength distribution. Notable differences in the corpora that could lead to this difference are that (1) our documents are multiple orders of magnitude longer than tweets; (2) our documents are written by legal specialists to protect companies from being sued and thus contain much more specific and less varied language than tweets.

% All co--occurrence network figures by year from 1997-2019 can be found in Supplementary Materials listed in Section \ref{section:online}.

\section{Discussion}
\label{section:discussion}

Our results can be summarized as follows: (1) privacy legislation correlates with turbulence and rates of change of privacy policies use of PII data type terms; (2) complexity of privacy policies decreases in years, which means that policies are becoming more regular over time; (3) sensitivity rises over time and shows spikes during the critical time when new privacy legislation is introduced. We find an increase in mentions of health and location data that is only partially explained by new legislation, which shows that new technologies with major privacy implications will be reflected in privacy policies even without changes in legislation. We also find evidence for insights as a mechanism for selling data without being classified as a data broker, which is becoming more common.

Looking at the basic properties of our dataset, we find that there is a lot of stability in the language and terms used in privacy policies over time and that the language is relatively consistent across companies, which makes this a promising dataset to extract more meaning from in future work.

The collection of PII data has become ubiquitous and poses a significant risk to data subjects. However, the lack of transparency about DBDPs who collect data on data subjects remains a large issue. Lack of transparency makes it difficult to assess the level of harm associated with harvesting PII data and hold DBDPs accountable for misuse of data. We hope this study humbly contributes critical insight into an important body of work related to privacy policy and data broker and data processor transparency. Much work remains to be done in this area, and we hope this analysis offers a small step toward bringing light to this important issue. 

In future work, we would like to investigate the flow of PII data types across sections of the privacy policies. With this future work, we hope to see what PII data types are present in the collect section of a privacy policy and compare that to the use and share section in privacy policies. We want to see what data is collected and then shared with third parties. 

% To potentially address the monetary harm and risk to data subjects, we would like to review tort law related to each of the PII data types and record the monetary amount of tort damages for each data type to estimate the value of the data types and the magnitude of these PII data collected over time.

% We would like to move from 1-gram detection to the extraction of concepts closer to data type definitions; word embeddings, for example, via BERT, are a promising approach, as is terminology extraction.

% \section{Citations and Bibliographies}

% \begin{verbatim}
%   \bibliographystyle{ACM-Reference-Format}
%   \bibliography{bibfile}
% \end{verbatim}

\begin{acks}
\begin{anonsuppress}
\section{Acknowledgments}
The authors would like to thank Jean-Gabriel Young, Laurent H\'ebert-Dufresne, and Nick Cheney for early discussions and feedback on this project. 

This work is supported by \grantsponsor{01} {MassMutual under the MassMutual Center of Excellence in Complex Systems and Data Science}{https://datascience.massmutual.com/}. \grantsponsor{02} {Alfred P. Sloan Foundation}{https://sloan.org/}. Any opinions, findings, and conclusions or recommendations expressed in this material are those of the author(s) and do not necessarily reflect the views of the aforementioned financial supporters. 
\end{anonsuppress}
\end{acks}

\bibliographystyle{ACM-Reference-Format}
\bibliography{BIBLIO}

% \section{Appendices}
\appendix

\section{Supplementary Materials}
 \label{appendix:I}

\subsection{Summary Statistics: Full corpus, PII data types corpus, network analysis}

In this section, we outline three tables with summary statistics: (1) summary statistics of the entire privacy policy corpus can be seen in Table \ref{tab:corpusdata}; (2) summary statistics of the PII privacy policy filtered corpus can be found in Table \ref{tab:piicorpusdata}; and summary statistics of the network properties of the co--occurrence networks can be found in Table \ref{tab:descdfdc2}.

 \begin{table*}[ht]
\small 
\centering
\caption{Overview of privacy policies corpus: the full corpus that is cleaned. Minimum description length (MDL) from the hSBM topic model, text description length (TDL) = total number of words x $(log_{2}$(unique words)), compression factor = MDL/TDL.}
\tabcolsep=0.19cm
\begin{tabular}{@{}*{7}{c}@{}}
\textbf{Year} & \textbf{No. words} & \textbf{No. unique words} & \textbf{No. policies full corp} & \textbf{MDL in bits} & \textbf{TDL} & \textbf{Compression Factor} \\
\hline
1997 & 4743 & 1148 & 9 & 8516.52666 & 48212.15355 & 0.1766468832\\
1998 & 80529 & 3901 & 144 & 191616.9197 & 960681.036 & 0.199459459\\
1999 & 413998 & 8059 & 602 & 991443.7243 & 5372197.486 & 0.1845508708\\
2000 & 2452053 & 19883 & 2886 & 5528846.65 & 35013472.48 & 0.1579062646\\
2001 & 4089391 & 26230 & 4442 & 9399703.064 & 60027885.0  & 0.156588943\\
2002 & 6009825 & 32451 & 6161 & 13659820.04 & 90063089.1 & 0.1516694594\\
2003 & 7558026 & 36958 & 7608 & 16829015.37 & 114682456.3 & 0.1467444622\\
2004 & 10051958 & 43265 & 9771 & 22154128.03 & 154809328.4 & 0.1431058984\\
2005 & 12485455 & 49056 & 11740 & 28084926.16 & 194550132.5 & 0.1443582988\\
2006 & 16721665 & 58317 & 15307 & 37314163.2 & 264731194.6 & 0.1409511382\\
2007 & 21688278 & 67080 & 19093 & 50251529.0 & 347741067.2 & 0.1445084683\\
2008 & 26488167 & 76818 & 23379 & 61419768.27 & 429880615.1 & 0.1428763385\\
2009 & 31196291 & 83596 & 27277 & 71748737.18 & 510095117.9 & 0.1406575649\\
2010 & 40154734 & 96377 & 34760 & 89700229.98 & 664817889.1 & 0.1349245131\\
2011 & 56365399 & 111383 & 46260 & 117489747.6 & 944975469.9 & 0.1243310026\\
2012 & 63254136 & 125455 & 50572 & 134453798.0 & 1071323311 & 0.1255025412\\
2013 & 72057214 & 138920 & 55591 & 152872571.3 & 1231017863 & 0.1241838773\\
2014 & 80135508 & 143880 & 59937 & 165640922.4 & 1373082386 & 0.1206343655\\
2015 & 88188734 & 148063 & 63820 & 179455129.4 & 1514716611 & 0.1184743919\\
2016 & 96232665 & 150585 & 66029 &  187656253.0 & 1655222869 & 0.1133721969\\
2017 & 94350462 & 146704 & 60343 & 179866323.9 & 1619294397 & 0.1110769754\\
2018 & 98640479 & 146546 & 49514 & 180949278.4 & 1692768675 & 0.1068954555\\ 
2019 & 114104999 & 163745 & 51791 & 208407776.4 & 1976423109  & 0.1054469438\\
\hline
\end{tabular}
\label{tab:corpusdata}
\end{table*}

\begin{table*}[ht]
\small
\centering
\caption{Overview of privacy policies PII data types corpus: includes a corpus filtered for just PII data type terms.}
\tabcolsep=0.19cm
\begin{tabular}{@{}*{4}{c}@{}}
\textbf{Year} & \textbf{No. words} & \textbf{Unique words} & \textbf{No. policies} \\
\hline
1997 &	25 & 13	 &	5 \\
1998 &		494 & 27 &		128\\
1999 &		2526 &		45	 &	546\\
2000 &		14522 &		64	 &	2671\\
2001 &		23182 &		71	 &	4156\\
2002 &		33963 &		76	 &	5723\\
2003 &		42670 &		76	 &	7059\\
2004 &		77980 &		78	 &	9079\\
2005 &		68965 &		78	 &	10912\\
2006 &		91992 &		78	 &	14321\\
2007 &		165744 &		80	 &	17940\\
2008 &		146335 &		80	 &	21983\\
2009 &		176552 &		84	 &	25719\\
2010 &		316706 &		85	 &	32996\\
2011 &		326307 &		85	 &	44209\\
2012 &		374757 &		86	 &	48449\\
2013 &		433677 &		84	 &	57515\\
2014 &		491032 &		83	 &	57822\\
2015 &		547066 &		84	 &	61707\\
2016 &		609593 &		84	 &	64064\\
2017 &		602959 &		86	 &	58672\\
2018 &		607413 &		86	 &	48337\\
2019 &		2972358 &		92	 &	50576\\
\hline
\end{tabular}
\label{tab:piicorpusdata}
\end{table*}

\begin{table*}[ht]
\small 
\centering
\caption{Network analysis on privacy policy text from 1997-2019. Metrics on the graph of co--occurrences of PII lexicon words for each year individually and also all years. All nodes with a degree less than one are filtered from the network (words that do not co--occur with other words). Here we explore the network measures of modularity, \cite{blondel2008fast} blocks, \cite{zhang2020statistical} average clustering coefficients, \cite{latapy2008main} average degree, and density.}
\tabcolsep=0.19cm
\begin{tabular}{@{}*{8}{c}@{}}
\textbf{Year} & \textbf{Nodes} & \textbf{Edges} & \textbf{Modularity (\# classes)} & \textbf{Blocks} & \textbf{Avg. clust. coeff.} & \textbf{Avg. Degree} & \textbf{Density} \\ \hline
1997 & 26 & 	192 & 	0.199 & 	3 & 	0.891 & 	14.7690 & 	0.591\\
1998 & 	65 & 	743 & 	0.096 & 	4 & 	0.798 & 	22.8620 & 	0.357\\
1999 & 	104 & 	1748 & 	0.029 & 	3 & 	0.828 & 	33.6150 & 	0.326\\
2000 & 	143 & 	3636 & 	0.027 & 	2 & 	0.837 & 	50.9930 & 	0.359\\
2001 & 	157 & 	4742 & 	0.025 & 	2 & 	0.833 & 	60.4080 & 	0.387\\
2002 & 	168 & 	5697 & 	0.031 & 	2 & 	0.830 & 	67.8210 & 	0.406\\
2003 & 	117 & 	6454 & 	0.032 & 	3 & 	0.824 & 	72.9270 & 	0.414\\
2004 & 	185 & 	7156 & 	0.029 & 	3 & 	0.830 & 	77.3620 & 	0.420\\
2005 & 	187 & 	7660 & 	0.029 & 	3 & 	0.829 & 	81.9250 & 	0.440\\
2006 & 	192 & 	8255 & 	0.028 & 	3 & 	0.835 & 	85.9900 & 	0.450\\
2007 & 	190 & 	8725 & 	0.025 & 	3 & 	0.833 & 	91.8320 & 	0.486\\
2008 & 	198 & 	9249 & 	0.024 & 	2 & 	0.837 & 	93.4240 & 	0.474\\
2009 & 	201 & 	9885 & 	0.024 & 	2 & 	0.835 & 	98.3580 & 	0.492\\
2010 & 	205 & 	10469 & 0.022 &  	3 & 	0.839 & 	102.1370 & 	0.501\\
2011 & 	216 & 	11913 & 0.041 & 	3 & 	0.843 & 	110.3060 & 	0.513\\
2012 & 	221 & 	12533 & 0.031 & 	3 & 	0.848 & 	113.4210 & 	0.516\\
2013 & 	220 & 	13045 & 0.031 & 	2 & 	0.850 & 	118.5910 & 	0.542\\
2014 & 	216 & 	13498 & 0.032 & 	2 & 	0.853 & 	124.9810 & 	0.581\\
2015 & 	219 & 	14127 & 0.032 & 	2 & 	0.855 & 	129.0140 & 	0.592\\
2016 & 	220 & 	14253 & 0.041 & 	2 & 	0.860 & 	129.5730 & 	0.592\\
2017 & 	221 & 	14670 & 0.041 & 	3 & 	0.860 & 	132.7600 & 	0.603\\
2018 & 	221 & 	15868 & 0.048 & 	2 & 	0.874 & 	143.6020 & 	0.653\\
2019 & 	240 & 	20847 & 0.050 & 	3 & 	0.894 & 	173.7250 & 	0.727\\
All Years & 251 & 	22334 & 0.039 & 2 & 	0.894 & 	177.9600 & 	0.712\\
\hline
\end{tabular}
\label{tab:descdfdc2}
\end{table*}

\subsection{Research Methods: Word level}

% \subsubsection{Rank shift:} To investigate the words used in the corpus and later to analyze PII data type frequency distributions between years and between industry types, we use the Allotaxonometer rank-shift divergence instrument \cite{dodds2020allotaxonometry, dodds2020probability}. This straightforward instrument allows us to compare normalized categorical frequency distributions of our PII data types and rank them in comparison with the entire corpus and comparison with other single years.

% We also look at the proportional word shift across policies over time. We use the \textit{Shifterator} tool developed by Gallagher et al. \cite{gallagher2021generalized}, which takes two corpora and visualizes a pairwise comparison between them through word shifts. Where \begin{math}p_{i}^{(1)}\end{math} is the relative frequency of word \textit{i} in the first corpus, and \begin{math}p_{i}^{(2)}\end{math} is its relative frequency in the second corpus, the proportion shift calculates their difference, where \begin{math}\delta p_{i} = p_{i}^{(2)} - p_{i}^{(1)}\end{math}. 

We first filter the corpus for both analyses to extract the terms in our PII data types lexicon. We then separate the privacy policy text by year, with 1997 as the first in the series and 2019 as the final corpus in the comparison. Frequency distribution is defined as the total count of a particular PII data type term and how many times it is used in a given corpus/total count of all words used in the given corpus. 

Rising looks at PII data types whose frequency increases quickly over time. We define rising words as words that rise in frequency by more than ten times in a 7-year period.

Falling looks at PII data types whose frequency drops quickly over time. Falling words are defined as words that fall in frequency by more than 15\% in a year.

Stability explores PII data types whose frequency distributions remain at similar frequencies over time with little change. Here stability is defined by the frequency distribution of words that change less than 2\% in frequency over a 20-year period. 

Emergence looks at new PII data types that appear in privacy policies over time. We define emergent words as words that occur at least 20 times in one year after not occurring the previous year. 

%We pair corpora in two-year intervals and investigate the word shift from 1997 to 2019.  We also compare single-year subcorpora to the entire corpus and investigate the word shifts. 
% The word shifts are represented through proportional word shift graphs that are paired together; in the graphs, if the word is relatively more common in the second corpus, then the word shift is positive, and if the word is more common in the first corpus, then the shift will be negative. Words are ranked by their relative difference between the first and second corpus. 

\subsection{Research Methods: Topic level}

\subsubsection{Stochastic block model topic model:} We conduct our topic modeling by first filtering all of the corpora to extract just the terms in our PII lexicon. We do this for the entire corpus, which includes privacy policies from 1997-2019, and then split it by year to compare individual years to the whole text. The filtering identifies words that are present in our PII lexicon and deletes all words that are not in the lexicon. The result leaves the word frequency of the PII terms and their relative placement in the text. We run the filtered text through a stochastic block model topic model to extract topic groups in the corpora. 

To understand what topic appears in the corpus, we will use the hSBM topic model implemented in Tiago Peixoto's Graph-tool. \cite{gerlach2018network} This Stochastic Block Model is a generative model that generalizes the Erd\H{o}s-Rényi model to have groups. It is a random Poisson graph model in that node degrees within any group are distributed according to a Poisson distribution. \cite{newman_networks_2018} The SBM is widely used in complex systems because it generates different network structures, e.g., core-periphery, community structure, and hub-and-spoke. Not only is the SBM flexible, but it is also extensible, as we can see below with the SBM Topic Model. The SBM topic model is a method that combines topic modeling and community detection through a word-document matrix (which is a bipartite network) that assumes communities represent topics. It uses maximum likelihood estimation to find a hierarchical clustering that fits the data.

A topic model is a method of extracting high-level information from textual data. In the hSBM topic model, we infer topics through the community structure of the document to word groupings. The hSBM model for topic modeling is a non-parametric symmetrical formulation that hierarchically clusters words and documents in a corpus. The tool divides words and documents into hierarchical groups with a list of constituent words per topic and the weight of the contribution of each of those words.

\subsubsection{Complexity measure:} From the graph-tool hSBM, \cite{gerlach2018network} we get the Minimum Description length (MDL) and convert the output of nats into bits (MDL bits = MDL nats x 1.4426950408889).  We then get the size of the original file, which we calculate as the total number of words in the corpus multiplied by the $log_{2}$ of the number of unique words in the corpus. \cite{hebert2022network} To compute the compression ratio, we divide the MDL in bits by the original file size in bits. We then compare the compression ratio in bits over time for the privacy policies. 

\subsection{Research Methods: Network level}

\subsubsection{co--occurrence networks:} To investigate what words are used in the corpus and later to see what data types are collected concurrently by DBDPs, we use co--occurrence network analysis to represent the use of relevant terms within the same text. We first filter all the corpora to extract the terms in our PII lexicon. co--occurrence networks utilize network analysis to represent the relationship between objects that co--occur in the same environment \cite{price2019symptoms, feicheng2014utilising, veling1999conceptual, fudolig2022sentiment}. To create a word co--occurrence network, we add an edge between words that appear in the same privacy policy. Edges start with a weight of one. If any two terms from our PII lexicon appear together in additional documents, we increase the weight based on the number of documents they co-occur.

To process our network visualizations, we remove all nodes with a degree less than one (words that do not co--occur with other words). Then networks are partitioned by modularity, which is signified by the color of the nodes. \cite{blondel2008fast} Finally, the node size ranges from 1 to 40 based on the weighted degree.

Most of our co--occurrence networks are partitioned into 2-4 classes. The co--occurrence networks seem to partition words into groups that can be generalized into two themes: PII data about a data subject's internet activity and PII data about the data subject's characteristics. An example of this can be seen in Figure \ref{fig:teaser}. In future work, we would like to take a closer look at the co--occurrence networks and investigate the dynamics of individual node attributes and how they change over time.

For visualization, we use a network backbone algorithm developed by Serrano et al. \cite{serrano2009extracting} to reduce the number of edges while preserving the relevant network structure. The algorithm works with a random network null model to compute the statistical significance of each link and drops links if their significance is below a certain threshold.

\section{Lexicon of Personally Identifiable Information (PII)}
\label{appendix:II}

'identifiers', 'alias', 'online identifier', 'internet protocol (ip) address', 'account name', 'social security number', 'passport number', 'customer records information', 'identification number', 'signature', 'electronic mail address', 'address', 'telephone number', 'protected health information', 'state identification card number', 'education', 'employment history', 'bank account number', 'face', 'financial information', 'records of personal property', 'products purchased', 'health condition', 'consuming histories', 'services purchased', 'eye color', 'retina scans', 'network activity', 'internet  activity', 'search history', 'geolocation', 'visual', 'thermal', 'olfactory', 'professional', 'medical condition', 'characteristics', 'aggregated data', 'predispositions', 'behavior', 'specific location', 'aptitudes', 'facial recognition', 'physiological', 'behavioral', 'audio', 'dna', 'iris', 'retina', 'hand', 'palm', 'vein patterns', 'voice recordings', 'minutiae template', 'health status', 'keystroke patterns', 'keystroke rhythms', 'gait rhythms', 'sleep', 'health', 'exercise', 'cross-context behavioral advertising', 'targeted advertising', 'dark pattern', 'personal information', 'racial', 'real name', 'account number', 'postal address', 'financial account number', 'internet protocol address', 'gender identity', 'email address', 'security question', 'color', 'religion', 'sex', 'sexual orientation', 'marital status', 'national origin', 'ancestry', 'genetic information', 'retaliation for reporting patient abuse in tax-supported institutions', 'age', 'religious dress', 'pregnancy', 'gender', 'childbirth', 'breastfeeding', 'mental characteristics', 'physical characteristics', 'hiv/aids', 'cancer', 'genetic characteristics', 'geolocation data', 'record of cancer', 'history of cancer', 'gender expression', 'abilities', 'mental condition', 'predict', 'biological', 'purchasing tendencies', 'aggregate consumer information', 'first name', 'voice', 'electronic network activity', 'biological characteristic', 'interaction with an advertisement', 'browsing history', 'employment', 'race', 'health records', 'citizenship', 'military or veteran status', 'medical identification number', 'access code', 'preferences', 'protected classifications', 'psychological trends', 'commercial information', 'medical information', 'attitudes', 'intelligence', 'name', 'driver's license', 'sensitive personal information', 'precise geolocation', 'locate', 'geographic area', 'radius', 'sensitive data', 'profiling', 'consumer's social security', 'driver's license', 'state identification card', 'account log-in', 'credit card', 'health insurance information', 'password', 'credentials allowing access to an account', 'combination', 'racial origin', 'ethnic origin', 'philosophical beliefs', 'union membership', 'text messages', 'genetic data', 'biometric data', 'personally identifiable information', 'security code', 'fingerprint', 'device identifier', 'ip address', 'cookies', 'beacons', 'pixel tags', 'customer number', 'unique pseudonym', 'telephone numbers', 'persistent identifier', 'probabilistic identifier', 'family', 'child', 'identifier template', 'de-identified data', 'health-care information', 'health-care provider', 'medicine', 'pharmacy', 'chiropractic', 'nursing', 'physical therapy', 'podiatry', 'dentistry', 'optometry', 'occupational therapy', 'healing arts', 'identified', 'identifiable individual', 'online identifier', 'personal data', 'products obtained', 'automated processing', 'request for pregnancy disability leave', 'analyze', 'economic situation', 'personal preferences', 'religious beliefs', 'reliability', 'location', 'movements', 'religious beliefs', 'physical health condition', 'diagnosis', 'credit card number', 'postal address', 'citizenship status', 'genetic', 'last name', 'mobile ad identifiers', 'health-care', 'patient', 'fingerprints', 'products considered', 'physical description', 'voiceprint', 'eye retinas', 'unique identifier', 'consuming tendencies', 'faceprint', 'driver's license number', 'services considered', 'global positioning system', 'latitude', 'longitude', 'coordinates', 'interests', 'financial account', 'user alias', 'irises','request for leave for an employee’s own serious health condition', 'condition', 'physical', 'diagnosis', 'insurance policy number', 'immigration status', 'known child', 'sex life', 'height', 'aids', 'medical diagnosis', 'religious grooming practices', 'identifiable individual', 'debit card', 'ethnic', 'origin', 'medical treatment', 'inferences', 'medical history', 'mental health', 'physical health', 'mental', 'biometric information', 'email content','physical representation','biological pattern', 'mother's maiden name', 'interaction with an internet website application', 'deoxyribonucleic acid', 'purchasing histories', 'disability', 'targeting of advertising', 'movements', 'hair color', 'digital representation', 'initials', 'specific geolocation data', 'driver authorization card number', 'identification card number', 'debit card number', 'health insurance identification number', 'user name', 'request for family care leave', 'date of birth', 'place of birth', 'unique biometric', 'human body', 'hiv', 'biometric', 'language', 'household', 'driver’s license number', 'government-issued identification number', 'driver license', 'nondriver state identification card number','individual taxpayer identification number', 'military identification card number', 'gait patterns', 'unique personal identifier', 'passwords', 'personal identification number', 'services obtained', 'wellness program', 'health promotion', 'disease prevention', 'health insurance policy number'

\section{Lexicon of Negation Words}
\label{appendix:III}
``don't'', ``never'', ``nothing'', ``nowhere'', ``noone'', ``none'', ``not'', ``hasn't'', ``hadn't'', ``can't'', ``couldn't'', ``shouldn't'', ``won't'', ``wouldn't'', ``don't'', ``doesn't'', ``didn't'', ``isn't'', ``aren't'', ``ain't'', ``in*'', ``un*'', ``dis*'', ``mal*''

\section{Topics}
\label{appendix:IV}
All topics we trace over time, as extracted from the full corpus, omitting words with weight $<0.001$:\\\\
\textbf{address-name-age:} address (0.5), name (0.241), age (0.097), password (0.085), physical (0.038), gender (0.019), passwords (0.013), locate (0.004), reliability (0.001)\\
\textbf{cookies:} cookies (1.0)\\
\textbf{preferences-analyze-movements:} preferences (0.71), analyze (0.252), movements (0.038)\\
\textbf{interests-exercise-profiling:} interests (0.664), exercise (0.292), profiling (0.045)\\
\textbf{employment-signature-height:} employment (0.789), signature (0.131), height (0.045), citizenship (0.036)\\
\textbf{professional-face:} professional (0.901), face (0.099)\\
\textbf{beacons:} beacons (1.0)\\
\textbf{location-geolocation-latitude:} location (0.926), geolocation (0.039), latitude (0.008), longitude (0.008), coordinates (0.007), inferences (0.006), fingerprint (0.004), radius (0.002)\\
\textbf{identified-identifiers-language:} identified (0.323), identifiers (0.223), language (0.207), combination (0.079), characteristics (0.056), audio (0.044), hand (0.029), color (0.013), visual (0.011), predict (0.011), attitudes (0.002)\\
\textbf{behavior-behavioral:} behavior (0.561), behavioral (0.439)\\
\textbf{family-ancestry-dna:} family (0.912), ancestry (0.056), dna (0.032)\\
\textbf{condition-household-pharmacy:} condition (0.439), household (0.271), pharmacy (0.083), cancer (0.061), medicine (0.054), diagnosis (0.047), aids (0.025), initials (0.016), dentistry (0.002), chiropractic (0.001)\\
\textbf{origin-ethnic-race:} origin (0.278), ethnic (0.155), race (0.155), sex (0.149), racial (0.107), religion (0.099), disability (0.057)\\
\textbf{health-patient-hiv:} health (0.951), patient (0.046), hiv (0.003)\\
\textbf{education:} education (1.0)\\
\textbf{child:} child (1.0)\\
\textbf{genetic-mental-biometric:} genetic (0.265), mental (0.261), biometric (0.238), physiological (0.075), sleep (0.06), pregnancy (0.054), fingerprints (0.021), biological (0.01), breastfeeding (0.005), predispositions (0.003), retina (0.002), olfactory (0.002), childbirth (0.002), voiceprint (0.002)\\
\textbf{voice-intelligence-palm:} voice (0.645), intelligence (0.257), palm (0.052), nursing (0.036), iris (0.01)

\begin{anonsuppress}
\section*{Author contributions statement}
J.L. conceived the study and developed the conceptual framework. J.L., P.M., and P.S., and P.D. developed the computational analysis and produced the results. All authors wrote and reviewed the manuscript.
\end{anonsuppress}
\begin{anonsuppress}
\section*{Additional information}
%To include, in this order: \textbf{Accession codes} (where applicable); 
\textbf{Competing interests} The authors declare no competing interests.\\ 

\section{Code availability statement}

Code associated with this project can be found on our \textcolor{blue}{\href{https://github.com/juniperlovato/privacypolicypaper}{Github repository}}.

\end{anonsuppress}

\end{document}